\documentclass[10pt,twocolumn,letterpaper]{article}

\usepackage{iccv}
\usepackage{times}
\usepackage{epsfig}
\usepackage{graphicx}
\usepackage{amsmath}
\usepackage{amssymb}
\usepackage{paralist}

\usepackage{mydefs}

\usepackage{booktabs}
\usepackage{makecell}
\usepackage{subcaption}
\usepackage{multirow}
\usepackage{pifont}
\newcommand{\cmark}{\ding{51}}

\usepackage{bbm}
\usepackage{mathtools}

\iccvfinalcopy

\usepackage[pagebackref=true,breaklinks=true,letterpaper=true,colorlinks,bookmarks=false]{hyperref}

\ificcvfinal
\newcommand{\od}{\textsc{Waymo Open Motion Dataset}}
\else
\newcommand{\od}{\textsc{Anonymized Dataset}}
\fi

\newcommand{\yes}[0]{\cmark}

\def\iccvPaperID{9324} %

\ificcvfinal\fi

\begin{document}

\title{Large Scale Interactive Motion Forecasting for Autonomous Driving : \\The \od{}}

\author{Scott Ettinger $^{1}$, Shuyang Cheng $^{1}$, Benjamin Caine $^{2}$, Chenxi Liu $^{1}$, Hang Zhao $^{1}$, Sabeek Pradhan $^{1}$, \\ Yuning Chai $^{1}$,  Ben Sapp $^{1}$, Charles Qi $^{1}$, Yin Zhou $^{1}$, Zoey Yang $^{1}$, Aurélien Chouard $^{1}$, Pei Sun $^{1}$, \\ Jiquan Ngiam $^{2}$, Vijay Vasudevan $^{2}$,  Alexander McCauley $^{1}$, Jonathon Shlens $^{2}$, Dragomir Anguelov $^{1}$ \\ $^{1}$ Waymo LLC, $^{2}$ Google Brain \\ }

\maketitle
\ificcvfinal\thispagestyle{empty}\fi

\begin{abstract}
 As autonomous driving systems mature, motion forecasting has received increasing attention as a critical requirement for planning. Of particular importance are interactive situations such as merges, unprotected turns, etc., where predicting individual object motion is not sufficient. Joint predictions of multiple objects are required for effective route planning. There has been a critical need for high-quality motion data that is rich in both interactions and annotation to develop motion planning models. In this work, we introduce the most diverse interactive motion dataset to our knowledge, and provide specific labels for interacting objects suitable for developing joint prediction models. With over 100,000 scenes, each 20 seconds long at 10 Hz, our new dataset contains more than 570 hours of unique data over 1750 km of roadways. It was collected by mining for interesting interactions between vehicles, pedestrians, and cyclists across six cities within the United States. We use a high-accuracy 3D auto-labeling system to generate high quality 3D bounding boxes for each road agent, and provide corresponding high definition 3D maps for each scene. Furthermore, we introduce a new set of metrics that provides a comprehensive evaluation of both single agent and joint agent interaction motion forecasting models. Finally, we provide strong baseline models for individual-agent prediction and joint-prediction. We hope that this new large-scale interactive motion dataset will provide new opportunities for advancing motion forecasting models. 
\end{abstract}
\section{Introduction}

\begin{figure}[t]
    \centering
    \begin{subfigure}{\linewidth}
    \centering
    \includegraphics[width=0.46\linewidth]{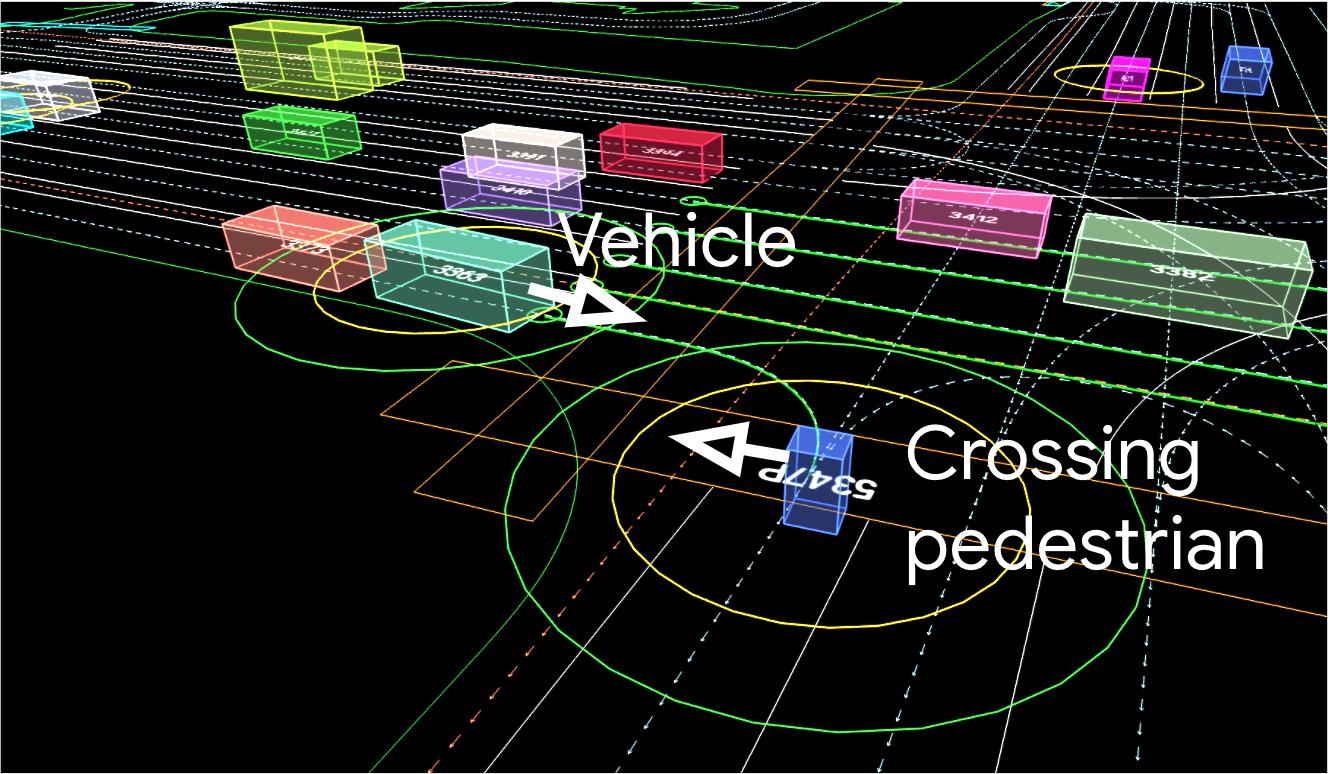}
    \hspace{0.05cm}
    \includegraphics[width=0.46\linewidth]{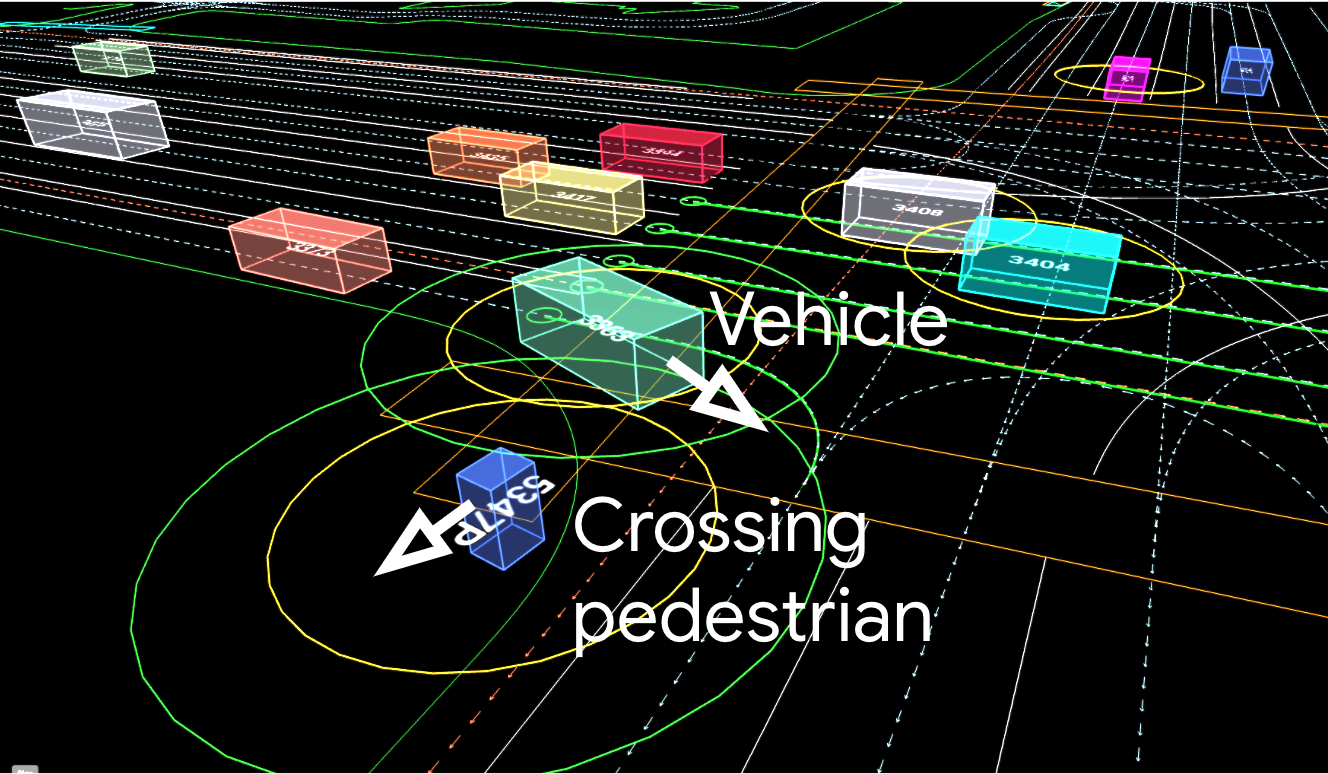}
    \caption{A vehicle waits for a pedestrian to fully cross the crosswalk before commencing a turn.}
    \vspace{0.1cm}
    \label{fig:yield}
    \end{subfigure}
    \begin{subfigure}{\linewidth}
    \centering
    \includegraphics[width=0.46\linewidth]{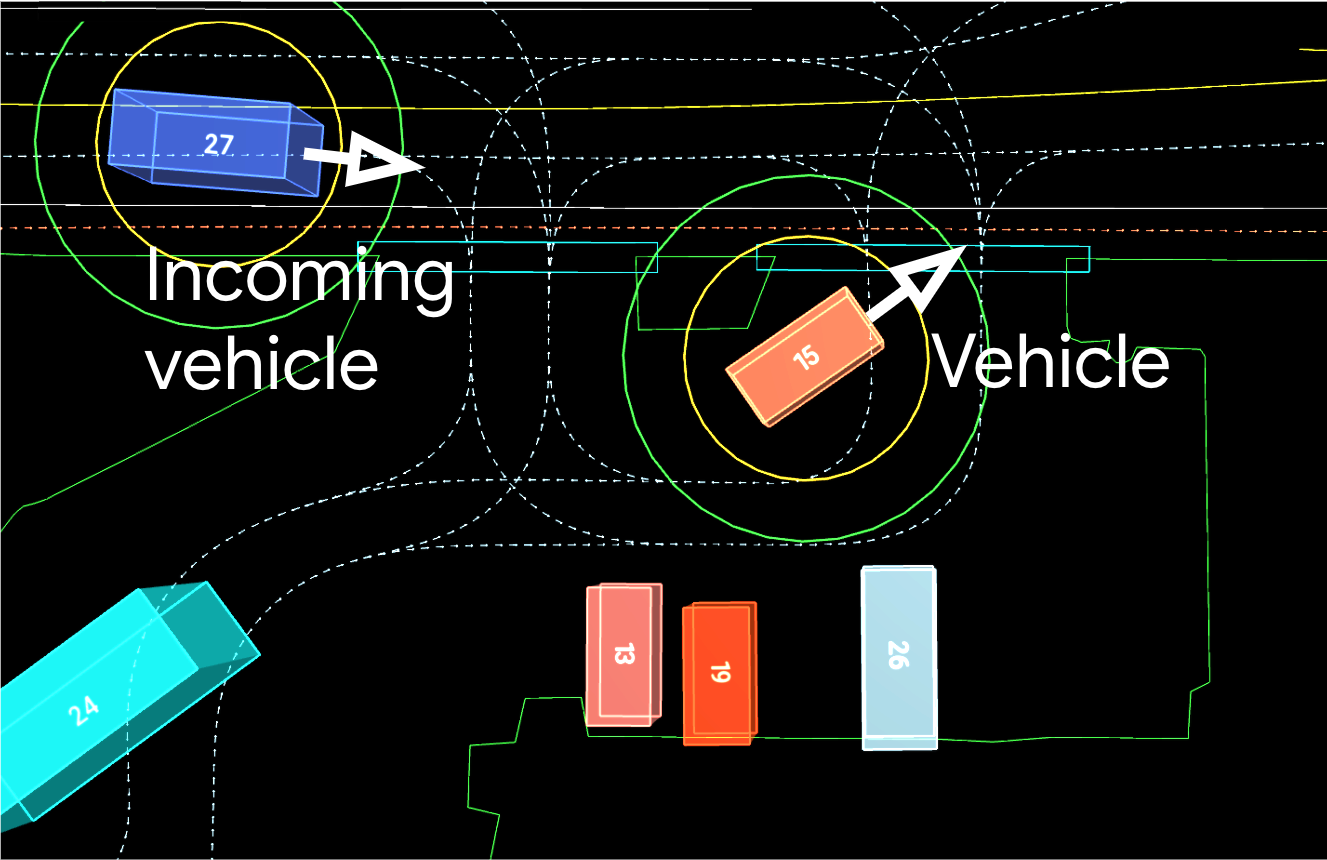}
    \hspace{0.05cm}
    \includegraphics[width=0.46\linewidth]{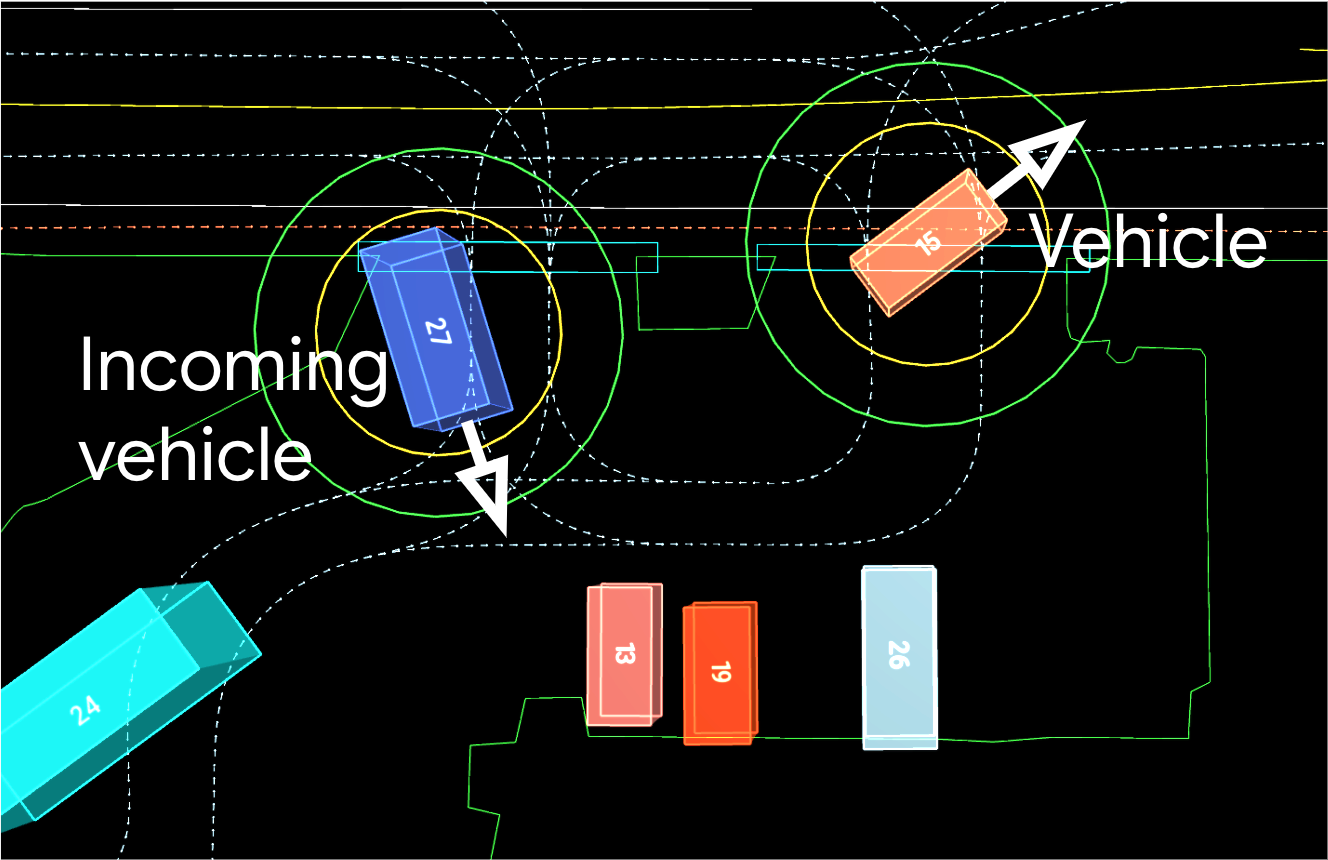}
    \caption{A vehicle accelerates onto the street only after the incoming vehicle turns.}
    \label{fig:accelerate}
    \end{subfigure}
    \caption{\textbf{Examples of interactions between agents in a scene in the \od{}.} Each example highlights how predicting the joint behavior of agents aids in predicting likely future scenarios. Solid and dashed lines indicate the road graph and associated lanes. Each numeral indicates a unique agent in the scene.}
    \vspace{-0.4cm}
\end{figure}

Motion forecasting has received increasing attention as a critical requirement for planning in autonomous driving systems \cite{chai2019multipath,gao2020vectornet,zhao2020tnt,kate_cbp,rhinehart2019precog,tang2019multiple}. Due to the complexity of scenes that autonomous systems need to safely handle, predicting object motion in the scene is a difficult task, suitable for machine learning models. Building effective motion forecasting models requires large amounts of high quality real world data. Creating a dataset for motion forecasting is complicated by the fact that the distribution of real world data is highly imbalanced \cite{caesar2020nuscenes,houston2020one,sun2020waymo_open,yu2020bdd100k}; in the common case, vehicles drive straight at a constant velocity. In order to develop effective models, a dataset must contain and measure performance on a wide range of behaviors and trajectory shapes for different object types that an autonomous system will encounter in operation.

We argue that critical situations (e.g., merges, lane changes, and unprotected turns) require the joint prediction of a set of multiple \textbf{interacting} objects, not just a single object. An example of a pedestrian and vehicle interacting is illustrated in Figure~\ref{fig:yield} where a vehicle waits for a pedestrian to fully cross the street before turning. In Figure~\ref{fig:accelerate}, the orange vehicle accelerates into the street only after ensuring the incoming blue vehicle's intention is to decelerate and turn off of the street. Most existing datasets have focused on single agent representation, but there has been considerably less work on interaction modeling at a large scale, which motivates this work.

The goal of this work is to provide a large scale, diverse dataset with specific annotations for interacting objects to promote the development of models to jointly predict interactive behaviors. In addition, we aim to supply object behaviors over a wide range of road geometries, and thus provide a large set of annotated interactions over a diverse set of locations. To generate such a set, we develop criteria for mining interactive behavior over a large corpus of driving data. We explicitly annotate groups of interacting objects in both training and validation/test data to enable development of models that jointly predict the motion of multiple agents as well as individual prediction models.

We aim to provide high quality object tracking data to reduce uncertainty due to perception noise. The cost of hand labeling a dataset of the required size is prohibitive. Instead we use a state-of-the-art automatic labeling system \cite{REF:3DAL2021} to provide high quality detection and tracking data of objects in the scenes. In contrast with many datasets which provide tracking from on-board autonomous systems, the off-board automatic labeling system provides higher accuracy as it is not constrained to run in real time. These high quality tracks allow us to focus on understanding the complexity of object behavior, rather than on dealing with perception noise.

Evaluation of interactive prediction models requires metrics formulated for joint predictions as motivated by recent work \cite{suo2021trafficsim, casas2020implicit, tang2019multiple, rhinehart2019precog}. In Section~\ref{sec:metrics}, we discuss existing work on generalizing metrics to the joint prediction case. We also propose a novel mean Average Precision (mAP) metric to capture the performance of models across different object types, prediction time scales, and trajectory shape buckets (\eg, u-turns, left turns). This method is inspired by metrics used in the object detection literature and overcomes limitations in currently adopted metrics. We discuss how this metric attempts to address issues with existing metrics.

We name our large-scale interactive motion dataset: \od{}. 
It will be made publicly available to the research community, and we hope it will provide new directions and opportunities in developing motion forecasting models.
We summarize the contributions of our work as follows:
\begin{compactitem}
    \item We release a large-scale dataset for motion forecasting research with specifically labeled interactive behaviors. The data is derived from high quality perception output across a large array of diverse scenes with rich annotations from multiple cities.
    \item We provide novel metrics for motion prediction analysis along with challenging benchmarks for both the marginal and joint prediction cases.
\end{compactitem}
\section{Related Work}

\begin{table}[t!]
    {
    \centering
    \scriptsize
    \begin{tabular}{c|ccccc}
    \toprule
         & Lyft & NuSc & Argo & Inter & Ours \\
         \midrule
         \# unique tracks & \,\,\,\,53.4\,m $^\mathsection$ & 4.3\,k & \,\,11.7\,m $^\ddagger$ & 40\,k & 7.64\,m \\
         Avg track length & 1.8\,s $^\mathsection$ & - & 2.48\,s $^\ddagger$ & 19.8\,s $^*$ & 7.04\,s $^{\dagger\dagger}$ \\
         Time horizon & 5\,s & 6\,s & 3\,s & 3\,s & 8\,s\\ 
         \# segments & 170k & 1k & 324k & - & 104k\\
         Segment duration & 25\,s & 20\,s & 5\,s & - & 20\,s \\
         Total time & 1118\,h & 5.5\,h & 320\,h & 16.5\,h\,$^*$ & 574\,h\\
         Unique roadways & 10\,km & - & 290\,km & - & 1750\,km$^{\dagger\dagger}$ \\
         \midrule
         Sampling rate & 10\,Hz & 2\,Hz & 10\,Hz & 10\,Hz & 10\,Hz\\
         \# cities covered & 1 & 2 & 2 & \,\,\,\,6\,$^*$ & 6 \\
         \# object types & 3 & \,\,\,\,1\,$^\dagger$ & \,\,\,\,1\,$^\ddagger$ & 1 & 3 \\
         Boxes & 2D & 3D & None & 2D & 3D \\
         3D maps &  &  & \yes &  & \yes \\
         Offline perception &  &  &  & \yes & \yes \\
         Interactions &  &  &  & \yes & \yes \\
         Traffic signal states & \yes &  &  &  & \yes \\
    \bottomrule
    \end{tabular}
    }
    \caption{\textbf{Comparison of popular behavior prediction and motion forecasting datasets.} Specifically, we compare Lyft Level 5 \cite{REF:lyftperception2019}, NuScenes \cite{caesar2020nuscenes}, Argoverse \cite{chang2019argoverse}, Interactions \cite{interactiondataset}, and our dataset across multiple dimensions. \# object types measures the number of types of objects to predict the motion trajectory. Dashed line "-" indicates that data is not available or not applicable. $^\mathsection$ Lyft Level 5 number of unique tracks and average track length are determined through private correspondence. $^\dagger$ nuScenes \cite{caesar2020nuscenes} provides annotations for 23 objects types (stationary vehicles are removed), but only vehicle is predicted. $^\ddagger$ Argoverse \cite{chang2019argoverse} provides annotations for 15 object types (Appendix B) but only vehicle is predicted. The number of unique tracks is determined through private correspondence. The average track length is estimated from data. $^*$ Interactions \cite{interactiondataset} gathered data from 4 countries including 6 cities (the last statistic is collected through personal communication) and the entire dataset is not divided into segments. The average track length is estimated from data. $^{\dagger\dagger}$ Our average track length is computed on the 20s segments of the training split. Our total unique roadway distance is calculated by hashing our autonomous vehicle poses as UTM coordinates into 25 meter voxels and counting the number of non-zero voxels.}
    \label{tab:datasets}
\end{table}

\paragraph{Motion forecasting datasets}

Several existing public datasets have been developed with the primary goal of motion forecasting in real-world urban driving environments, compared in Table~\ref{tab:datasets}.  The datasets vary in size measured in number of scenes, total time, total miles, number of tracked objects, and number of distinct time segments. While Lyft Level 5~\cite{REF:lyftperception2019} has the most hours of data and NuScenes~\cite{caesar2020nuscenes} has rich object taxonomy, they were not collected to capture a wide diversity of complex and interactive driving scenarios. Argoverse~\cite{chang2019argoverse} was collected for interesting behaviors by biasing sampling towards certain observed behaviors (\eg, lane changes, turns) and road features (\eg, intersections). The INTERACTION dataset~\cite{interactiondataset} manually selected a small set of specific driving locations (e.g., roundabouts), and times of day (e.g., rush hour) to obtain a dataset with high interaction complexity.  We explain our own methodology for collecting interactions in Section~\ref{sec:scenario_mining}.

Another salient dataset attribute is the time horizon for prediction.  Our dataset's forecasting horizon is 8 seconds into the future, considerably longer than others (3 or 5 seconds), as we believe that long term forecasting is necessary for safe and human-like planning, and is intrinsically more difficult.  Finally, most datasets are auto-labeled with industry-grade, \textit{onboard} 3D perception stacks, employing LiDAR's, cameras, and/or radar, and provided as-is with noisy state estimates and tracking errors.  One exception is the INTERACTION dataset~\cite{interactiondataset} which collects data from drone footage, which is then post-processed offline with detection, tracking and track smoothing.  We also put considerable effort into creating high quality state estimates and 3D tracks by employing an offboard 3D detection and tracking pipeline, as discussed in Section~\ref{sec:perception}.

We consider perception datasets (e.g., KITTI~\cite{geiger2013vision}, Waymo Open Dataset~\cite{sun2020waymo_open}) outside of the scope of this discussion as they do not contain enough motion data to build sufficiently complex models.  We also note there are a host of other motion forecasting datasets which, while popular, are orders of magnitude smaller, have $O(10)$ unique locations, and/or are not focused on driving environment, for example the Stanford Drone Dataset~\cite{robicquet2016learning}, NGSIM~\cite{coifman2017ngsim}, ETH~\cite{pellegrini2009you}, UCY~\cite{lerner2007crowds}, Town Center~\cite{benfold2011stable}.

\paragraph{Jointly consistent multi-agent forecasting} Most existing models output independent future distributions per object in a scene, \eg~\cite{alahi2016social,buhet2020plop,casas2018intentnet,casas2020spagnn,chai2019multipath,cui2019multimodal,cui2020dkm,gao2020vectornet,hong2019rules,lee2017desire,liang2020laneGCN,phan2019covernet,zhao2020tnt}.  This is encouraged by the popular metrics, which only measure quality on a per-object level, and by datasets that only require predicting one agent per scene.  An important note is that these methods {\em do} model interactions between objects to achieve better performance, but explicitly modeling joint futures is much less common.  There are a few exceptions which model jointly-consistent futures: Precog~\cite{rhinehart2019precog} and MFP~\cite{tang2019multiple} employ models which roll out trajectory samples timestep-by-timestep, where each agent's next step sample conditions on all other agents' current and past steps.  In contrast, ILVM~\cite{casas2020implicit} (also used by TrafficSim~\cite{suo2021trafficsim}), samples from a latent variable from which multiple steps of future joint samples from all agents are decoded, without explicit conditioning on each step of rollout.  These works all measure a stricter version of distance error metrics, reporting the per-agent error of the best {\em joint} configuration.  It is important to note that none of the datasets in Table~\ref{tab:datasets} provide such joint metrics in their release, in contrast to our~\od.
\section{Dataset}

The dataset provides high quality object tracks generated using an offboard perception system (described in Section \ref{sec:perception}) along with both static and dynamic map features to provide context for the road environment. Object track states are sampled at 10Hz. Each state includes the object's bounding box (3D center point, heading, length, width, and height), and the object's velocity vector. Due to sensor range or occlusion, measurements of an object's state may not exist at some time steps. A valid flag is provided to indicate which time steps have valid measurements. Map data is provided as a set of polylines and polygons created from curves sampled at a resolution of 0.5 meters. Static map feature types include lane centers, lane boundary lines, road edges, stop signs, crosswalks, and speed bumps. Traffic signal states and the lanes they control are included. In addition to the geometry data, map features also contain additional data specific to each feature type e.g. lane boundaries have a field to indicate if they are a broken white boundary, a double yellow boundary, etc. 

Starting with 20 second segments that are specifically mined from interactions as described in \ref{sec:scenario_mining}, we create 9.1 second (91 steps at 10Hz) scenes, splitting the data into a 70\% training, 15\% validation, and 15\% test set. We derive two versions of the validation and test sets which we refer to as the standard and interactive versions. The standard validation and test sets provide up to 8 objects to predict in each scene. Selection is biased to require objects that do not follow a constant velocity model or straight paths. The interactive versions of the validation and test sets focus on the interactive portion of the segment and require only the 2 mined interactive objects to be predicted. The original 20 second segments are also provided for research requiring longer time frames.
 
\subsection {Mining for interesting scenarios}
\label{sec:scenario_mining}

We mine for interesting scenarios by first hand-crafting semantic predicates involving agents' relationships---\eg,  ``agent $A$ changed lanes at time $t$'', and ``agents $A$ and $B$ crossed paths with a time gap $t$ and relative heading difference $\theta$''.
These predicates can be composed to retrieve more complex queries in an efficient SQL and relational database framework on an overall data corpus orders of magnitude larger than the resulting curated \od.

With this framework, we specifically mined for the following pairwise interaction scenarios: merges, lane changes, unprotected turns, intersection left turns, intersection right turns, pedestrian-vehicle interactions, cyclist-vehicle interactions, interactions with close proximity, and interactions with high accelerations. The pair of interacting objects is annotated within the dataset in each scenario, and the interaction happens close to the 10s mark of the 20s clip.

\subsection {Dataset statistics}

\begin{figure}[t]
    \centering
    \includegraphics[width=0.9\linewidth]{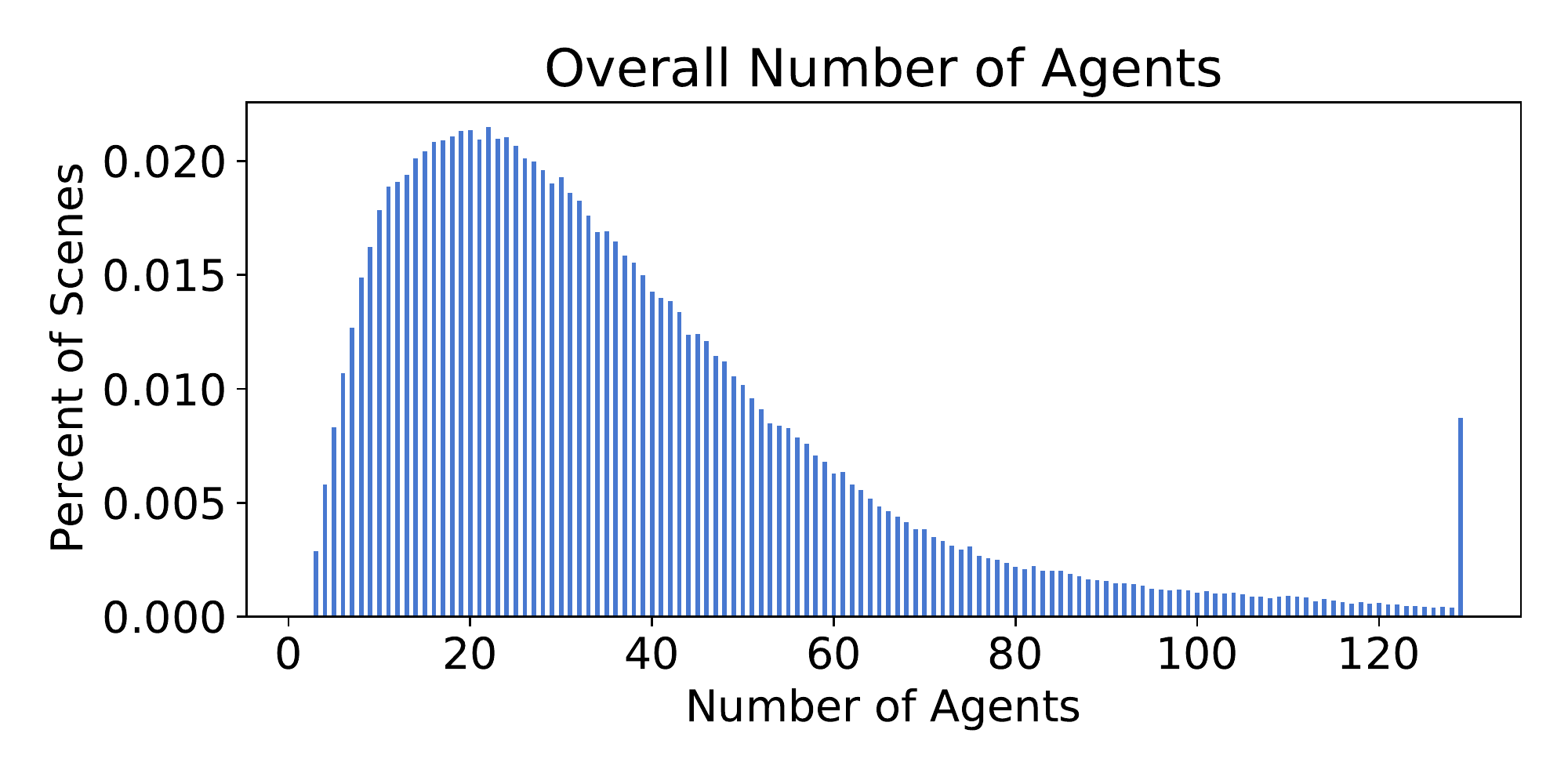}
	\includegraphics[width=0.8\linewidth]{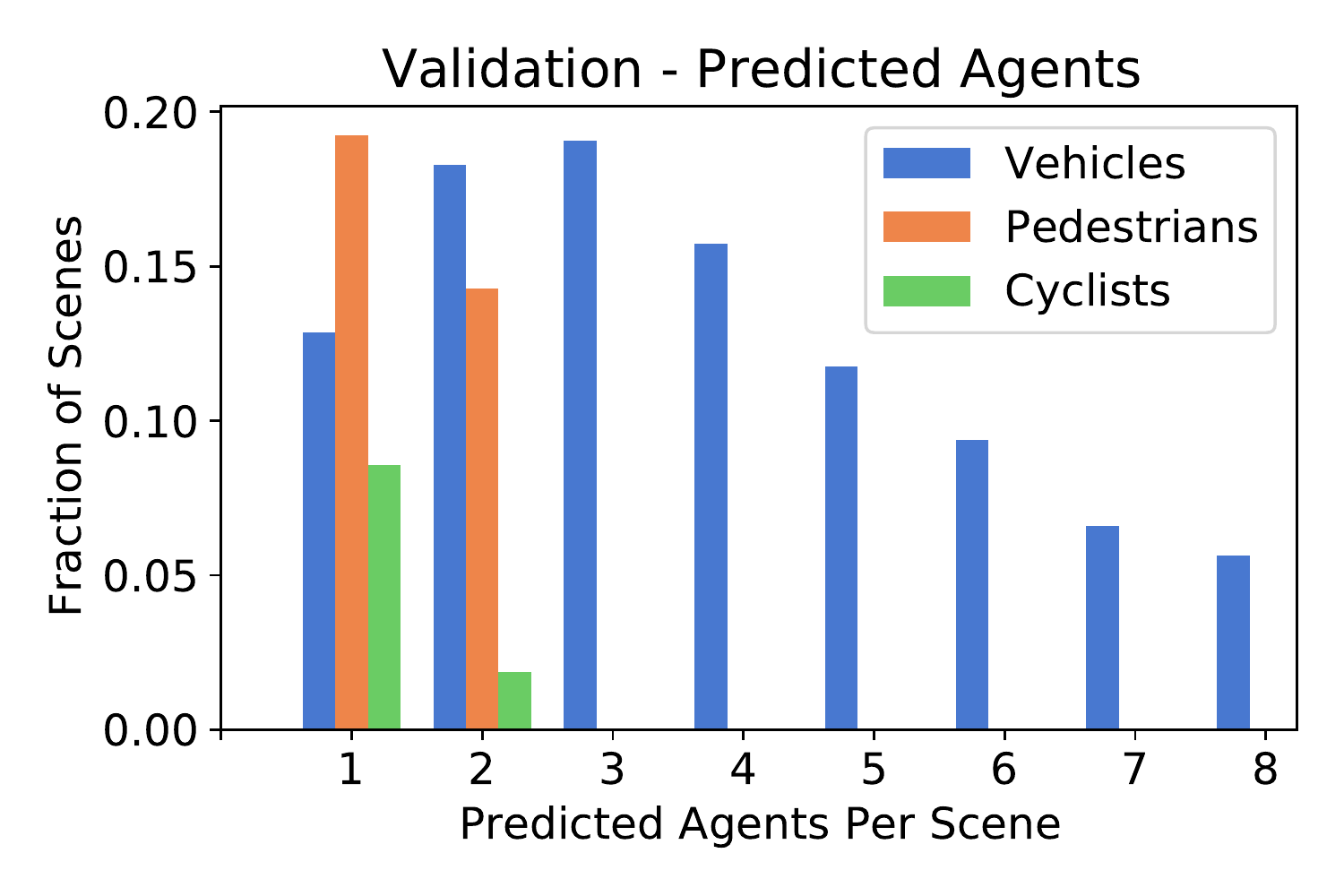}
    \vspace{-0.4cm}
    \caption{\textbf{Our dataset contains many agents including pedestrians and cyclists.} Top: 46\% of scenes have more than 32 agents, and 11\% of scenes have more than 64 agents. Bottom: In the standard validation set, 33.5\% of scenes require at least one pedestrian to be predicted, and 10.4\% of scenes require at least one cyclist to be predicted.}
    \label{fig:agents}
\end{figure}

In contrast with many existing datasets that provide a limited number of agents per scene or agent types, we provide more diverse scenes in terms of the number of agents and types of agents, reflecting many complicated real world driving scenarios like city driving and busy intersections. We show the distribution of number of agents per scene (Figure \ref{fig:agents}, top). All scenes have at least one vehicle, 57\% of scenes have at least one pedestrian (with 20\% having four or more), and 16\% of scenes have at least one cyclist.

In addition to accurately predicting the motion of other vehicles, to safely drive, an autonomous vehicle must also accurately predict the motion of other road agents like pedestrians and cyclists. To support this, our dataset contains rich interactions between vehicles, pedestrians, and cyclists, and the users of this dataset must be able to accurately predict the trajectories of \textit{all three classes}, which is not the case in previous datasets \cite{chang2019argoverse, caesar2020nuscenes, interactiondataset}. We show the frequency of scenes in which we ask the model to predict each class in the validation set (Figure \ref{fig:agents}, bottom). Notably, 38.3\% of scenes in the validation set require the model to predict more than one type of agent (e.g. a vehicle and a pedestrian or cyclist), and 4.9\% of scenes require a model to predict trajectories for all three classes. Finally, in the interactive validation set, where we task the model with predicting the joint future trajectories of two interacting agents, 77.5\% of scenes involve two interacting vehicles, 14.9\% of scenes involve a vehicle interacting with a pedestrian, and 7.6\% of scenes involve a vehicle interacting with a cyclist.

\begin{figure}[t]
    \centering
    \includegraphics[width=0.95\linewidth]{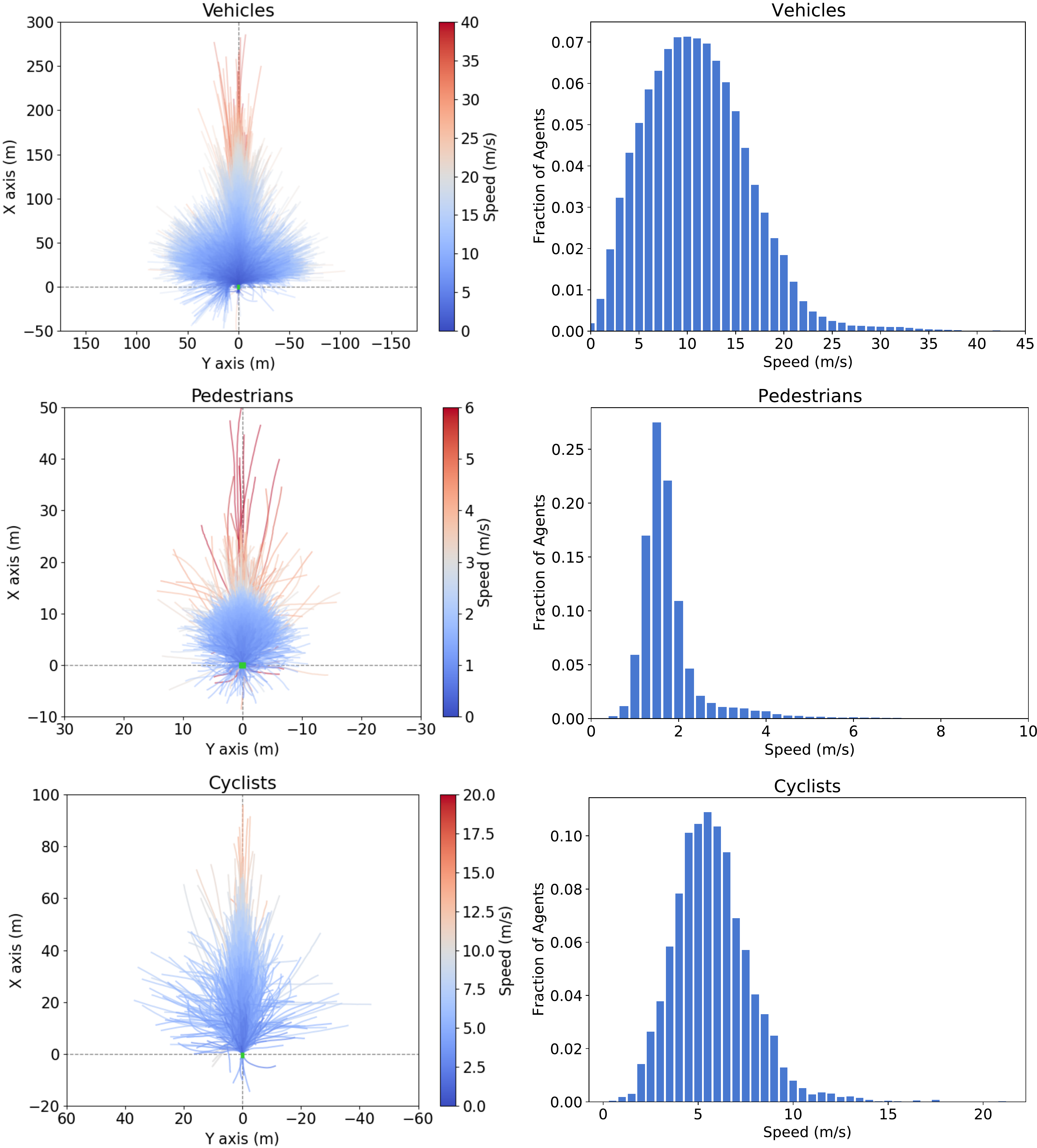}
    \caption{\textbf{Agents selected to be predicted have diverse trajectories.} Left: Ground truth trajectory of each predicted agent in a frame of reference where all agents start at the origin with heading pointing along the positive $X$ axis (\textit{pointing up}). Right: Distribution of maximum speeds achieved by all of the agents along their 9 second trajectory. Plots depict variety in trajectory shapes and speed profiles.
    }
    \label{fig:agent_trajectories}
\end{figure}

Finally, a motion forecasting dataset should contain diverse scenarios, trajectories, and agent interactions. Table \ref{tab:datasets} shows that we gather data across a large range of roadways. Figure \ref{fig:agent_trajectories} visualizes the future ground-truth trajectories and maximum speeds of agents we task the models with predicting. These agents represent a wide range of trajectory shapes, speeds, and behaviors, which we believe accurately captures the many different behavioral modes for each class.

\subsection{Offboard perception system}
\label{sec:perception}
Modern motion forecasting systems require a large amount of training data to imitate human maneuvers in complex real-world scenarios. Recently released datasets for motion forecasting \cite{chang2019argoverse,houston2020one,caesar2020nuscenes} are orders of magnitude larger than popular 3D perception datasets \cite{caesar2020nuscenes,REF:lyftperception2019,sun2020waymo_open,geiger2013vision}. However, manually annotating datasets at such large scales not only incurs exorbitant cost but it also takes tremendous amount of time~\cite{REF:3DAL2021,REF:yang2021auto4d}. 
Constrained by the high cost, most existing motion forecasting datasets \cite{chang2019argoverse,houston2020one} directly employ onboard perception output as groundtruth for trajectory prediction. 
But limited by the onboard perception system performance, such annotated 3D objects tracks may have a high degree of state estimation error, lack temporal kinematic consistency or under-/over-segment tracks.

In this work, we aim to alleviate the perception quality bottleneck in existing motion datasets captured by autonomous vehicles and propose using the recently introduced offboard algorithms~\cite{REF:3DAL2021,REF:yang2021auto4d} to automatically generate high-quality motion labels, allowing motion forecasting algorithms to focus on the subtle dynamics and interactions of agents instead of overcoming the noise generated by a constrained, onboard perception system. Compared to the onboard counterpart, offboard perception has two major advantages: 1) it can afford much more powerful models running on the ample computational resources; and 2) it can maximally aggregate complementary information from different views by exploiting the full point cloud sequence including both history and future. Thanks to those advantages, the offboard perception system has shown superior perception accuracy compared to onboard detectors~\cite{REF:3DAL2021} and we have further validated its quality in Section~\ref{sec:box_quality}.

The offboard perception system~\cite{REF:3DAL2021} employed contains three steps: (1) 3D object detector generates object proposals from each LiDAR frame. (2) Multi-object tracker links detected objects throughout the LiDAR sequence. (3) For each object, an object-centric refinement network processes the tracked object boxes and its point clouds across all frames in the track, and outputs temporally consistent and accurate 3D bounding boxes of the object in each frame.

\section {Metrics}
\label{sec:metrics}

To measure the accuracy of motion predictions we use a suite of five metrics, which we extend to handle joint predictions over multiple agents as proposed by a few related works~\cite{tang2019multiple, casas2020implicit, rhinehart2019precog}.  Several common metrics report a minimum error within a trajectory set; when generalized, the joint metric analog constrains the minimum over the best joint configuration of trajectories from a group of agents. 

We report standard trajectory-set distance error metrics minADE, minFDE, and Miss Rate (MR), with a custom definition of a match explained below.  We also report overlap rate (OR) to measure frequency of predicted tracks' extents overlapping with others'.  Finally, inspired by the detection literature, we propose an Average Precision (AP) metric according to the defined MR to measure the precision and recall performance of models across different confidence values.  We then account for imbalanced data by reporting mean AP (mAP) over different semantic trajectory motion types.

For each sample $\bme$, a model makes $K$ possibly joint predictions $S_k, k \in 1 \ldots K$. Each $S_k$ contains a scalar confidence $c_k$, and a trajectory $\bms^k = \{s_{a,t}\}_{t=1:T,a=1:A}$ for $T$ future time steps for $A$ agents. Similarly, the ground truth is denoted as $\bshat = \{\hat{s}_{a,t}\}$.    The individual object prediction task becomes a special case of this formulation where each joint prediction contains only a single agent $A=1$.  

\PAR{minADE.} The minimum Average Displacement Error computes the L2 norm between $\bshat$ and the closest joint prediction: $\frac{1}{TA} \min_k \sum_a \sum_t ||\hat{s}_{a,t} - s_{a,t}^{k}||_2$.

\PAR{minFDE.} The minimum Final Displacement Error is equivalent to evaluating the minADE at a single time step $T$: $ \frac{1}{A} \min_k \sum_a ||\hat{s}_{a,T} - s_{a,T}^{k}||_2$

\PAR{Overlap rate (OR).} The overlap rate is computed by taking the highest confidence joint prediction from each multi-modal joint prediction. If any of the A agents in the jointly predicted trajectories overlap at any time with any other objects that were visible at the prediction time step (compared at each time step up to T) or with any of the jointly predicted trajectories, it is considered a single overlap. The overlap rate is computed as the total number of overlaps divided by the total number of multi-modal joint predictions. See the supplementary material for details.  The overlap is calculated using box intersection, with box extents taken as the current time step's estimates, and heading inferred from consecutive waypoint position differences.

\PAR{Miss rate (MR).} A binary match/miss indicator function $\textsc{IsMatch}(\hat{s}_t, s_t)$  is assigned to each sample waypoint at a time $t$. The average over the dataset creates the miss rate at that time step.
Our dataset asks to predict an 8-seconds trajectory on agents with varying speed profiles. Therefore, a single distance threshold to determine $\textsc{IsMatch}$ is insufficient: we want a stricter criteria for slower moving and closer-in-time predictions, and also different criteria for lateral deviation (\eg wrong lane) versus longitudinal (\eg wrong speed profile).  
For a particular joint configuration, a miss is assigned for time $t$ if any of the trajectories don't match their ground truth trajectory:  $ \text{MR}_t = \min_k \lor_a \lnot IsMatch(\hat{s}_t, s_{a,t}^k)$. We implement $\textsc{IsMatch}$ with separate lateral and longitudinal thresholds, which scale as a clamped linear function of future time and velocity. See the supplementary material for details.

\PAR{Mean average precision (mAP).} The Average Precision computes the area under the precision-recall curve by applying confidence score thresholds $c_k$ across a validation set, and using the definition of Miss Rate above to define true positives, false positives, \etc. 
Consistent with object detection mAP metrics~\cite{lin2014mscoco}, only one true positive is allowed for each object and is assigned to the highest confidence prediction.

Further inspired by object detection literature~\cite{Everingham2009ThePV}, we seek an overall metric balanced over semantic buckets, some of which may be much more infrequent (\eg, u-turns), so report the mean AP over different driving behaviors.  The final mAP metric averages over eight different ground truth trajectory shapes: straight, straight-left, straight-right, left, right, left u-turn, right u-turn, and stationary.

\begin{table*}[t]
    \centering
    \footnotesize
    \bgroup
    \def\arraystretch{1.05}%
    \begin{tabular}{clccc|ccc|ccc|ccc}
    \toprule
    & & & & & \multicolumn{3}{c|}{Vehicle} & \multicolumn{3}{c|}{Pedestrian} & \multicolumn{3}{c}{Cyclist}\\
    Set & Model & \texttt{rg} & \texttt{ts} & \texttt{hi} & minADE $\downarrow$ & MR $\downarrow$ & mAP $\uparrow$ & minADE $\downarrow$ & MR $\downarrow$ & mAP $\uparrow$ & minADE $\downarrow$ & MR $\downarrow$ & mAP $\uparrow$ \\
    \midrule
    \multirow{7}{*}{\makecell{Standard\\Validation}} & Const. Vel. & & & & 11.0 & 0.95 & 0.02 & 1.55 & 0.60 & 0.07 & 4.17 & 0.82 & 0.02 \\
    \cmidrule{2-14}
    & \multirow{6}{*}{LSTM} & & & & 2.63 & 0.67 & 0.07 & 0.73 & 0.22 & 0.15 & 1.86 & 0.60 & 0.07 \\
    & & & & \cmark & 1.67 & 0.40 & 0.16 & 0.74 & 0.18 & 0.18 & 1.50 & 0.40 & 0.12 \\
    & & \cmark & &  & 1.54 & 0.32 & 0.19 & 0.66 & 0.14 & \textbf{0.23} & 1.36 & 0.31 & 0.17 \\
    & & \cmark & & \cmark & 1.36 & 0.26 & 0.22 & \textbf{0.63} & 0.14 & \textbf{0.23} & 1.29 & 0.30 & 0.18 \\
    & & \cmark & \cmark & & 1.52 & 0.31 & 0.18 & 0.65 & 0.15 & 0.20 & 1.34 & 0.33 & 0.15 \\
    & & \cmark & \cmark & \cmark & \textbf{1.34} & \textbf{0.25} & \textbf{0.23}  & \textbf{0.63} & \textbf{0.13} & \textbf{0.23} & \textbf{1.26} & \textbf{0.29} & \textbf{0.21} \\
    \midrule
    \midrule
    \multirow{2}{*}{\makecell{Standard\\Test}} & Const. Vel. & & & & 11.0 & 0.95 & 0.02 & 1.58 & 0.60 & 0.06 & 4.12 & 0.83 & 0.03 \\
    \cmidrule{2-14}
    & LSTM & \cmark & \cmark & \cmark & \textbf{1.34} & \textbf{0.24} & \textbf{0.24} & \textbf{0.64} & \textbf{0.13} & \textbf{0.22} & \textbf{1.29} & \textbf{0.28} & \textbf{0.20} \\
    \bottomrule
    \end{tabular}%
    \egroup
    \caption{\textbf{\underline{Marginal} metrics on the \underline{standard} validation and test set.} All metrics computed at 8s. \texttt{rg} stands for road graph information. \texttt{ts} stands for traffic signal states information. \texttt{hi} stands for high-order interactions between agents' features. The constant velocity baseline employs $K = 1$ predicted trajectories; all other models employ $K = 6$.}
    \label{tab:baselines_regular}
\end{table*}

\begin{table*}[t]
    \centering
    \footnotesize
    \bgroup
    \def\arraystretch{1.05}%
    \begin{tabular}{clccc|ccc|ccc|ccc}
    \toprule
    & & & & & \multicolumn{3}{c|}{Vehicle} & \multicolumn{3}{c|}{Pedestrian} & \multicolumn{3}{c}{Cyclist}\\
    Set & Model & \texttt{rg} & \texttt{ts} & \texttt{hi} & minADE $\downarrow$ & MR $\downarrow$ & mAP $\uparrow$ & minADE $\downarrow$ & MR $\downarrow$ & mAP $\uparrow$ & minADE $\downarrow$ & MR $\downarrow$ & mAP $\uparrow$ \\
    \midrule
    \multirow{7}{*}{\makecell{Interactive\\Validation}} & Const. Vel. & & & & 10.3 & 0.98 & 0.00 & 3.62 & 1.00 & 0.00 & 6.35 & 1.00 & 0.00  \\
    \cmidrule{2-14}
    & \multirow{6}{*}{LSTM} & & & & 4.16 & 0.88 & 0.01 & 2.45 & 0.93 & 0.02 & 4.00 & 0.98 & 0.00 \\
    & & & & \cmark & 2.89 & 0.75 & 0.06 & \textbf{2.22} & 0.93 & 0.01 & 3.75 & 0.94 & 0.01 \\
    & & \cmark & & & 2.94 & 0.75 & 0.04 & 2.39 & \textbf{0.86} & 0.06 & 3.30 & 0.88 & 0.02 \\
    & & \cmark & & \cmark & 2.45 & \textbf{0.66} & 0.06 & \textbf{2.22} & \textbf{0.86} & 0.03 & \textbf{3.02} & \textbf{0.83} & \textbf{0.03} \\
    & & \cmark & \cmark & & 2.92 & 0.75 & 0.04 & 2.69 & 0.93 & \textbf{0.10} & 3.24 & 0.89 & 0.01 \\
    & & \cmark & \cmark & \cmark & \textbf{2.42} & \textbf{0.66} & \textbf{0.08} & 2.73 & 1.00 & 0.00 & 3.16 & \textbf{0.83} & 0.01 \\
    \midrule
    \midrule
    \multirow{2}{*}{\makecell{Interactive\\Test}} & Const. Vel. & & & & 10.3 & 0.98 & 0.01 & 4.56 & 1.00 & \textbf{0.00} & 6.21 & 1.00 & 0.00 \\
    \cmidrule{2-14}
    & LSTM & \cmark & \cmark & \cmark & \textbf{2.46} & \textbf{0.67} &\textbf{0.08} & \textbf{2.47} & \textbf{0.89} & \textbf{0.00} & \textbf{2.96} & \textbf{0.89} & \textbf{0.01} \\
    \bottomrule
    \end{tabular}%
    \egroup
    \caption{\textbf{\underline{Joint} metrics on the \underline{interactive} validation and test set.} See Table \ref{tab:baselines_regular} for abbreviations and details. Note that these metrics indicate that the interactive split is systematically more challenging.}
    \label{tab:baselines_interactive}
\end{table*}

\section{Experiments}
In this section, we evaluate various baseline models on the {\od} to investigate the importance of rich map annotations (\eg 3D road graph, traffic signal states), interaction context, and joint modeling (Section~\ref{ssec: baseline}). We then compare the standard validation and interactive validation datasets on conditional behavior prediction metrics to show that the interactive validation dataset is both more challenging and more interactive (Section~\ref{sec:interactivity}). Furthermore, we show that our offboard perception system achieves a similar accuracy and perception noise reduction to human labels (Section~\ref{sec:box_quality}). Finally, to provide insight on the performance measurement of motion prediction tasks, we empirically analyze minADE vs. mAP on their ability to reflect the quality of confidence score calibration (Section~\ref{sec:compare_metrics}).

\subsection{Baseline model performances} \label{ssec: baseline}

In this section, we evaluate several baseline models on the proposed dataset.
First, we consider a {\it Constant Velocity} model in which we assume the agent will maintain its velocity at the \texttt{current} timestamp for all \texttt{future} steps. Second, we consider a family of deep-learned models using various encoders, with a base architecture of an LSTM to encode a 1-second history of observed state~\cite{hochreiter1997long, alahi2016social}; this includes agents' positions, velocity, and 3D bounding boxes. In order to measure the importance of particular additional features, we selectively provide additional information:
\begin{compactitem}
    \item Road graph (\texttt{rg}): Encode the 3D map information with polylines following \cite{gao2020vectornet}.
    \item Traffic signals (\texttt{ts}): Encode the traffic signal states with an LSTM encoder as an additional feature. 
    \item High-order interactions (\texttt{hi}): Model the high-order interactions between agents with a global interaction graph following \cite{gao2020vectornet}.
\end{compactitem}
In experiments, combinations of these encodings are concatenated together to create an embedding per-agent, in agent-centered coordinates.  We decode K=6 trajectories for output using another MLP with min-of-k loss~\cite{cui2019multimodal,thiede2019analyzing}. See the supplementary material for details.

In Table~\ref{tab:baselines_regular} and \ref{tab:baselines_interactive}, we report the marginal metrics on the standard validation/test set and joint metrics on the interactive validation/test set, respectively.
Specifically, minADE, miss rate, and mAP at 8s are chosen to be the representatives, and we break down the metrics across 3 object types. 
The constant velocity model performs quite poorly, e.g., achieving double digit minADE on vehicles.
This shows that our dataset contains nontrivial trajectories. 

We then investigate the importance of encoding 3D map information, traffic signal states, and high-order interactions between agents.
Intuitively, they should all benefit motion forecasting, and this is indeed supported by the experimental results.
For example, on the standard validation set (Table~\ref{tab:baselines_regular}) for vehicle trajectory prediction, minADE improves from 2.63 to 1.34 and mAP improves from 0.07 to 0.23 when incrementally adding more information in this order.
The same trend holds for pedestrian and cyclist as well.

We only evaluate joint metrics on the interactive sets.
Since making joint predictions is a relatively new practice, there are no mature, established baselines.
In Table~\ref{tab:baselines_interactive}, we reuse the models trained to make $K$ marginal predictions; but when evaluating on the 2 interactive agents, we select the top $K$ among the $K^2$ possibilities based on the product of predicted probabilities, as described in~\cite{casas2020implicit}. 
The overall low performance in Table~\ref{tab:baselines_interactive} can be attributed to at least 3 factors: the higher difficulty level of the mined interactive agents; the requirement to make good predictions for \emph{both} agents as dictated by the joint version of the metrics; the fact that the predictions are post-hoc manipulations rather than the result of true joint training.

We have argued the importance of jointly predicting interactive behaviors. 
In Table~\ref{tab:joint} we provide direct comparison between a base LSTM (without \texttt{rg}, \texttt{ts}, or \texttt{hi}) trained to make marginal or joint predictions for the 2 interactive agents.
In converting the marginal model to making joint predictions, the neural features for the 2 interactive agents are concatenated with each other to provide the minimal necessary context; the sum of their individual distances to the ground truth (while matching the pairs of trajectories jointly) are used for training; the confidence score are jointly predicted for each pair of trajectories to ensure consistency.
When evaluated on the interactive set using joint metrics, this joint model performs favorably against its marginal counterpart. 
We hope this preliminary experiment can motivate further development of joint models on our dataset, especially the interactive set.

\begin{table}[t]
    \centering
    \footnotesize
    \begin{tabular}{l|ccc|ccc}
    \toprule
    & \multicolumn{3}{c|}{Vehicle minADE $\downarrow$} & \multicolumn{3}{c}{Vehicle mAP $\uparrow$} \\
    Model & 3s & 5s & 8s & 3s & 5s & 8s \\
    \midrule
        Marginal & {\bf 0.65} & 1.66 & 4.16 & 0.08 & \bf{0.07} & 0.01 \\
        Joint & \bf{0.65} & \bf{1.59} & \bf{3.81} & \bf{0.10} & 0.06 & \bf{0.03} \\
    \bottomrule
    \end{tabular}
    \caption{\textbf{Joint modeling is advantageous on interactive agents.} Numbers are from the interactive validation set.}
    \label{tab:joint}
    \vspace{-0.4cm}
\end{table}

\subsection{Quantifying interactivity}
\label{sec:interactivity}

Following \cite{kate_cbp}, we use Conditional Behavior Prediction (CBP) to quantify the interactivity in our dataset. \cite{kate_cbp} introduces a model that can produce either unconditional predictions or predictions conditioned on a ``query trajectory'' for one of the agents in the scene. If two agents are not interacting, then one's actions have no effect on the other, so knowledge of that agent's future should not change predictions for the other agent. Thus, \cite{kate_cbp} defines the \emph{degree of influence} agent A has on agent B as the KL divergence between the unconditional predictions for B and the predictions for B conditioned on A's ground truth future trajectory.

\begin{figure}[h]
    \centering
    \includegraphics[width=0.8\linewidth]{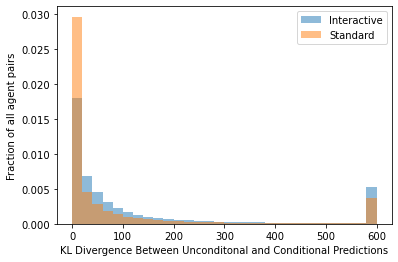}
    \caption{\textbf{The interactive split sees much larger improvements from conditional prediction.} Each element in the histogram is one pair of query agent/target agent, and the x axis shows the KL divergence between the unconditional predictions on the target agent and the predictions for the target agent conditioned on the query agent's ground truth future. Note that both plots are normalized to the total number of agent pairs.}
    \label{fig:cbp_kld_hist}
\end{figure}

We apply this framework to our interactive and standard validation datasets, computing the KL divergence between unconditional and conditional predictions for every query agent/target agent pair in the dataset. We find that the KL divergences are much larger in the interactive validation dataset than in the standard validation dataset. In particular, 73\% of agent pairs in the interactive dataset have KL divergences greater than 10, and 45\% have KL divergences greater than 50; in the standard dataset, these numbers are 48\% and 28\% respectively. Figure \ref{fig:cbp_kld_hist} presents a full histogram of the KL divergences between unconditional and conditional prediction for each agent pair. Conditioning on a query agent's future trajectories makes little difference in the standard validation dataset but a large difference in the interactive validation dataset, providing evidence that the interactive dataset contains more cases where multiple agents are interacting with and influencing each other. For details on the CBP model, see the supplementary material.

\begin{figure}[h]
    \centering
    \includegraphics[width=\linewidth]{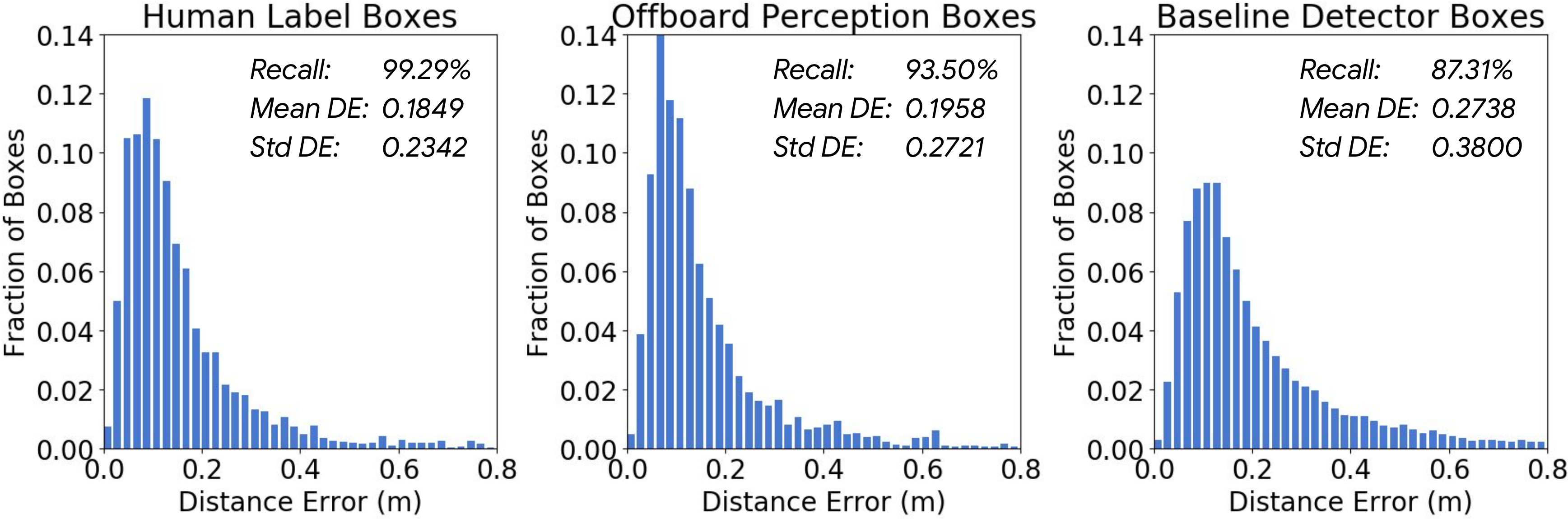}
    \caption{\textbf{Distance error statistics of vehicle bounding boxes.} We compare three sets of vehicle bounding boxes with the Waymo Open Dataset (WOD) ground truth boxes on the 5 selected run segments from the val set. The statistics include the histogram of distance errors (capped at 0.8m), the box recall (using a 3D IoU threshold of 0.03), mean distance error and standard deviation (std) of the distance error. Only boxes with at least one point inside are considered. Note that the DE from different boxes are not directly comparable as the recalls are different.}
    \label{fig:box_distance_error_hist}
\end{figure}

\subsection{Analysis of perception data quality}
\label{sec:box_quality}
In this section, we study the quality of our offboard perception system and compare them with two alternatives -- human labels and baseline detector boxes.
Following~\cite{REF:3DAL2021}, we conduct a study on the same five validation set run segments from the Waymo Open Dataset (WOD) re-labeled by extra three independent human labelers. With the duplicate human labels, we can analyze the human label consistency to understand the ``background noise'' in label accuracy. Instead of comparing detection results in average precision \cite{REF:3DAL2021}, we evaluate the box distance errors (DE) in meters by comparing to the original WOD ground truth boxes.

Figure~\ref{fig:box_distance_error_hist} shows that offboard perception achieves an accuracy and distance error distribution similar to human labels. We also show the distance errors of boxes obtained from a baseline detector (Multi-view Fusion~\cite{zhou2020end}) with a Kalman filter-based tracker (the same tracker used in the offboard perception). Using the baseline (onboard) detector leads to a significantly higher mean distance error -- this increased perception noise indicates a higher lower-bound minADE that a behavior model can achieve.

\subsection{Comparing mAP with minADE}
\label{sec:compare_metrics}
While minADE is widely adopted for performance measurement in motion forecasting tasks~\cite{chang2019argoverse,chai2019multipath,gao2020vectornet,zhao2020tnt}, it fails to measure the quality of confidence score calibration in the trajectory prediction. In contrast, the mAP metric described in Section~\ref{sec:metrics} provides a measurement of the quality of the confidence score calibration by design. In this section, we perform an analysis of minADE vs. mAP with increasing numbers of predictions at different time steps to show that minADE does not provide a full picture of the model performance while mAP provides more insight.

\begin{figure}[t]
    \centering
    \includegraphics[width=0.85\linewidth]{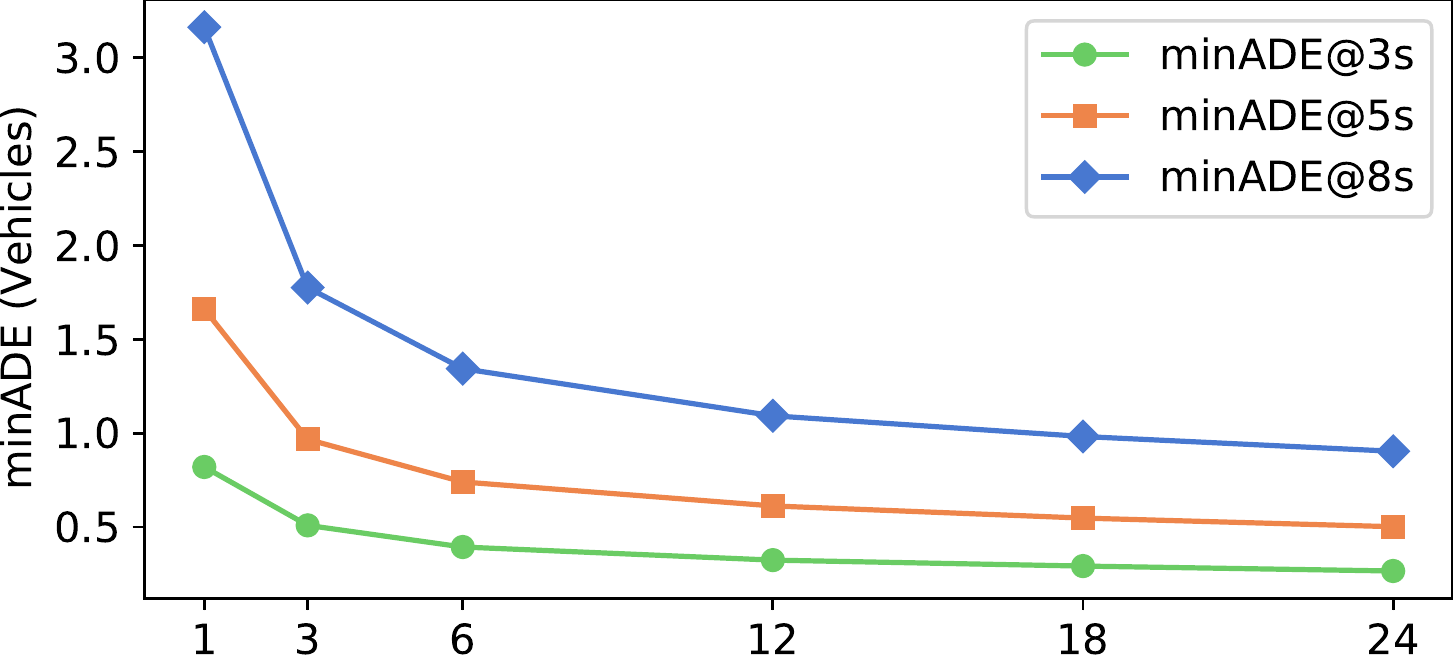}
    \includegraphics[width=0.85\linewidth]{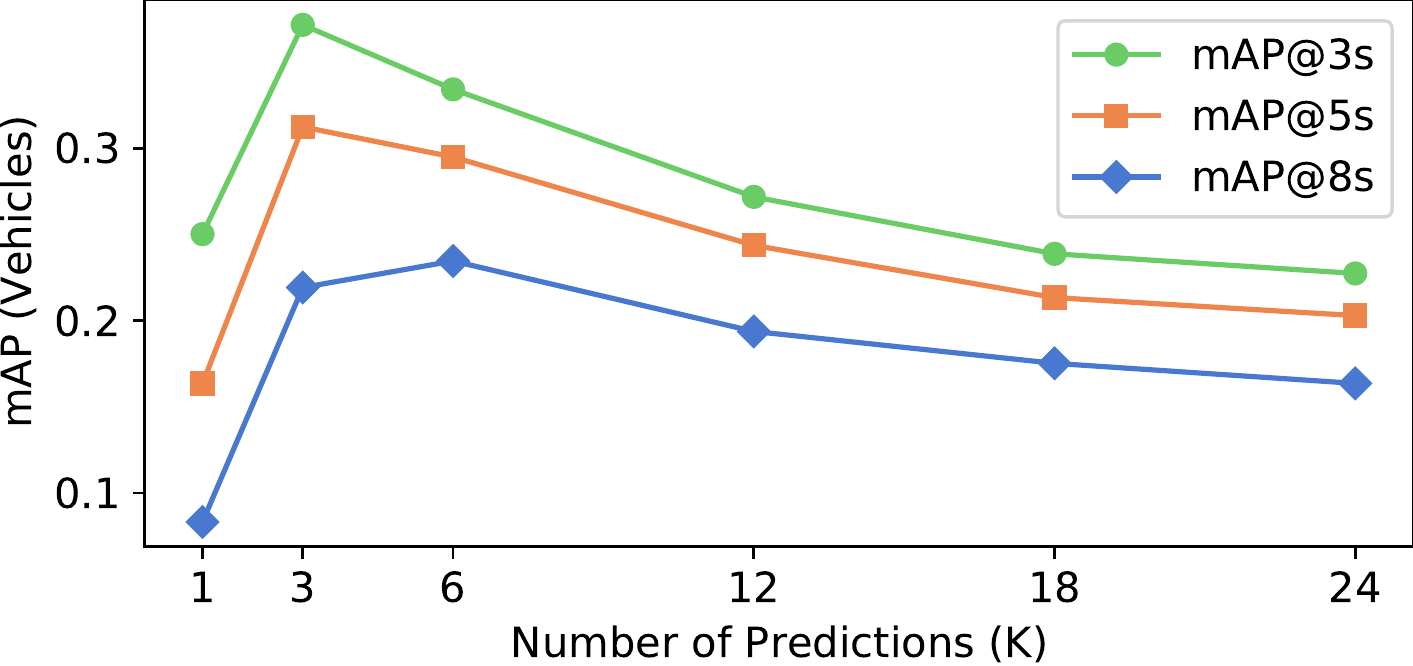}
    \caption{\textbf{Comparison of minADE and mAP across increasing numbers of predictions.} Using the best LSTM baseline model in Section~\ref{ssec: baseline}, the minADE (top) artificially improves as one allows for increasing numbers of predictions. Conversely, the mAP (bottom) saturates as the model must produce high quality confidence estimates in addition to accurate trajectories.}
    \label{fig:minADE_mAP_vs_k}
\end{figure}

As shown in Figure~\ref{fig:minADE_mAP_vs_k}, minADE artificially improves as the number of predictions increase, while the mAP value peaks at 3 predictions for 3s and 5s, and at 6 predictions for 8s. The minADE scores may improve so long as any of the predictions are good regardless of their confidence score. In contrast, mAP penalizes high confidence false positive predictions and does not continue to improve with the number of predictions. Precision-recall curves for these experiments are shown in the supplementary material.

\section{Discussion}
In this work we release the \od{}, a large-scale motion forecasting dataset containing data mined for interactive behaviors across a diverse set of road geometries from multiple cities. The data comes with rich 3D object state and HD map information.  Object tracks are generated with a state-of-the-art offboard automatic labeling system which is significantly higher fidelity than typical onboard 3D perception stacks. For evaluation we outline a set of metrics for both per-agent and joint trajectory predictions, including a novel mAP metric to measure precision-recall performance in a balanced way across semantic driving behavior buckets.  We provide baseline models for both individual and interactive prediction tasks, which we hope provides great opportunities for advancing motion forecasting research.

\ificcvfinal
\section*{Acknowledgements}
We thank Paul Hempstead, David Margines, Dietmar Ebner, Peter Pawlowski, Balakrishnan Varadarajan, Avikalp Srivastava, Zhifeng Chen, and Rebecca Roelofs for their comments and suggestions. Additionally, we thank the larger Google Brain team and Waymo Research teams for their support.
\fi

\clearpage

{\small
\bibliographystyle{ieee_fullname}
\bibliography{main}

\begin{thebibliography}{10}\itemsep=-1pt

\bibitem{alahi2016social}
Alexandre Alahi, Kratarth Goel, Vignesh Ramanathan, Alexandre Robicquet, Li
  Fei-Fei, and Silvio Savarese.
\newblock Social lstm: Human trajectory prediction in crowded spaces.
\newblock In {\em Proceedings of the IEEE conference on computer vision and
  pattern recognition}, pages 961--971, 2016.

\bibitem{benfold2011stable}
Ben Benfold and Ian Reid.
\newblock Stable multi-target tracking in real-time surveillance video.
\newblock In {\em CVPR 2011}, pages 3457--3464. IEEE, 2011.

\bibitem{buhet2020plop}
Thibault Buhet, Emilie Wirbel, and Xavier Perrotton.
\newblock Plop: Probabilistic polynomial objects trajectory planning for
  autonomous driving.
\newblock {\em arXiv preprint arXiv:2003.08744}, 2020.

\bibitem{caesar2020nuscenes}
Holger Caesar, Varun Bankiti, Alex~H Lang, Sourabh Vora, Venice~Erin Liong,
  Qiang Xu, Anush Krishnan, Yu Pan, Giancarlo Baldan, and Oscar Beijbom.
\newblock nuscenes: A multimodal dataset for autonomous driving.
\newblock In {\em Proceedings of the IEEE/CVF Conference on Computer Vision and
  Pattern Recognition}, pages 11621--11631, 2020.

\bibitem{casas2020spagnn}
Sergio Casas, Cole Gulino, Renjie Liao, and Raquel Urtasun.
\newblock Spagnn: Spatially-aware graph neural networks for relational behavior
  forecasting from sensor data.
\newblock In {\em 2020 {IEEE} International Conference on Robotics and
  Automation, {ICRA} 2020, Paris, France, May 31 - August 31, 2020}, pages
  9491--9497. {IEEE}, 2020.

\bibitem{casas2020implicit}
Sergio Casas, Cole Gulino, Simon Suo, Katie Luo, Renjie Liao, and Raquel
  Urtasun.
\newblock Implicit latent variable model for scene-consistent motion
  forecasting.
\newblock In {\em Proceedings of the European Conference on Computer Vision
  (ECCV)}. Springer, 2020.

\bibitem{casas2018intentnet}
Sergio Casas, Wenjie Luo, and Raquel Urtasun.
\newblock Intentnet: Learning to predict intention from raw sensor data.
\newblock In {\em Conference on Robot Learning}, pages 947--956. PMLR, 2018.

\bibitem{chai2019multipath}
Yuning Chai, Benjamin Sapp, Mayank Bansal, and Dragomir Anguelov.
\newblock Multipath: Multiple probabilistic anchor trajectory hypotheses for
  behavior prediction.
\newblock {\em arXiv preprint arXiv:1910.05449}, 2019.

\bibitem{chang2019argoverse}
Ming-Fang Chang, John Lambert, Patsorn Sangkloy, Jagjeet Singh, Slawomir Bak,
  Andrew Hartnett, De Wang, Peter Carr, Simon Lucey, Deva Ramanan, et~al.
\newblock Argoverse: 3d tracking and forecasting with rich maps.
\newblock In {\em Proceedings of the IEEE/CVF Conference on Computer Vision and
  Pattern Recognition}, pages 8748--8757, 2019.

\bibitem{coifman2017ngsim}
Benjamin Coifman and Lizhe Li.
\newblock A critical evaluation of the next generation simulation (ngsim)
  vehicle trajectory dataset.
\newblock {\em Transportation Research Part B: Methodological}, 105:362--377,
  2017.

\bibitem{cui2020dkm}
Henggang Cui, Thi Nguyen, Fang-Chieh Chou, Tsung-Han Lin, Jeff Schneider, David
  Bradley, and Nemanja Djuric.
\newblock Deep kinematic models for kinematically feasible vehicle trajectory
  predictions.
\newblock In {\em 2020 IEEE International Conference on Robotics and Automation
  (ICRA)}, pages 10563--10569. IEEE, 2020.

\bibitem{cui2019multimodal}
Henggang Cui, Vladan Radosavljevic, Fang-Chieh Chou, Tsung-Han Lin, Thi Nguyen,
  Tzu-Kuo Huang, Jeff Schneider, and Nemanja Djuric.
\newblock Multimodal trajectory predictions for autonomous driving using deep
  convolutional networks.
\newblock In {\em 2019 International Conference on Robotics and Automation
  (ICRA)}, pages 2090--2096. IEEE, 2019.

\bibitem{Everingham2009ThePV}
M. Everingham, L. Gool, C.~K. Williams, J. Winn, and Andrew Zisserman.
\newblock The pascal visual object classes (voc) challenge.
\newblock {\em International Journal of Computer Vision}, 88:303--338, 2009.

\bibitem{gao2020vectornet}
Jiyang Gao, Chen Sun, Hang Zhao, Yi Shen, Dragomir Anguelov, Congcong Li, and
  Cordelia Schmid.
\newblock {VectorNet}: Encoding hd maps and agent dynamics from vectorized
  representation.
\newblock In {\em {CVPR}}, 2020.

\bibitem{geiger2013vision}
Andreas Geiger, Philip Lenz, Christoph Stiller, and Raquel Urtasun.
\newblock Vision meets robotics: The kitti dataset.
\newblock {\em The International Journal of Robotics Research},
  32(11):1231--1237, 2013.

\bibitem{hochreiter1997long}
Sepp Hochreiter and J{\"u}rgen Schmidhuber.
\newblock Long short-term memory.
\newblock {\em Neural computation}, 9(8):1735--1780, 1997.

\bibitem{hong2019rules}
Joey Hong, Benjamin Sapp, and James Philbin.
\newblock Rules of the road: Predicting driving behavior with a convolutional
  model of semantic interactions.
\newblock In {\em {CVPR}}, 2019.

\bibitem{houston2020one}
John Houston, Guido Zuidhof, Luca Bergamini, Yawei Ye, Ashesh Jain, Sammy
  Omari, Vladimir Iglovikov, and Peter Ondruska.
\newblock One thousand and one hours: Self-driving motion prediction dataset.
\newblock {\em arXiv preprint arXiv:2006.14480}, 2020.

\bibitem{REF:lyftperception2019}
R. Kesten, M. Usman, J. Houston, T. Pandya, K. Nadhamuni, A. Ferreira, M. Yuan,
  B. Low, A. Jain, P. Ondruska, S. Omari, S. Shah, A. Kulkarni, A. Kazakova, C.
  Tao, L. Platinsky, W. Jiang, and V. Shet.
\newblock Lyft level 5 perception dataset 2020.
\newblock \url{https://level5.lyft.com/dataset/}, 2019.

\bibitem{lee2017desire}
Namhoon Lee, Wongun Choi, Paul Vernaza, Christopher~B Choy, Philip~HS Torr, and
  Manmohan Chandraker.
\newblock Desire: Distant future prediction in dynamic scenes with interacting
  agents.
\newblock In {\em Proceedings of the IEEE Conference on Computer Vision and
  Pattern Recognition}, pages 336--345, 2017.

\bibitem{lerner2007crowds}
Alon Lerner, Yiorgos Chrysanthou, and Dani Lischinski.
\newblock Crowds by example.
\newblock In {\em Computer graphics forum}, volume~26, pages 655--664. Wiley
  Online Library, 2007.

\bibitem{liang2020laneGCN}
Ming Liang, Bin Yang, Rui Hu, Yun Chen, Renjie Liao, Song Feng, and Raquel
  Urtasun.
\newblock Learning lane graph representations for motion forecasting.
\newblock {\em arXiv preprint arXiv:2007.13732}, 2020.

\bibitem{lin2014mscoco}
Tsung-Yi Lin, Michael Maire, Serge Belongie, James Hays, Pietro Perona, Deva
  Ramanan, Piotr Doll{\'a}r, and C~Lawrence Zitnick.
\newblock Microsoft coco: Common objects in context.
\newblock In {\em European conference on computer vision}, pages 740--755.
  Springer, 2014.

\bibitem{pellegrini2009you}
Stefano Pellegrini, Andreas Ess, Konrad Schindler, and Luc Van~Gool.
\newblock You'll never walk alone: Modeling social behavior for multi-target
  tracking.
\newblock In {\em 2009 IEEE 12th International Conference on Computer Vision},
  pages 261--268. IEEE, 2009.

\bibitem{phan2019covernet}
Tung Phan-Minh, Elena~Corina Grigore, Freddy~A Boulton, Oscar Beijbom, and
  Eric~M Wolff.
\newblock {CoverNet}: Multimodal behavior prediction using trajectory sets.
\newblock {\em arXiv:1911.10298}, 2019.

\bibitem{REF:3DAL2021}
Charles~R. Qi, Yin Zhou, Mahyar Najibi, Pei Sun, Khoa Vo, Boyang Deng, and
  Dragomir Anguelov.
\newblock Offboard 3d object detection from point cloud sequences, 2021.

\bibitem{rhinehart2018r2p2}
Nicholas Rhinehart, Kris~M Kitani, and Paul Vernaza.
\newblock R2p2: A reparameterized pushforward policy for diverse, precise
  generative path forecasting.
\newblock In {\em Proceedings of the European Conference on Computer Vision
  (ECCV)}, pages 772--788, 2018.

\bibitem{rhinehart2019precog}
Nicholas Rhinehart, Rowan McAllister, Kris Kitani, and Sergey Levine.
\newblock Precog: Prediction conditioned on goals in visual multi-agent
  settings.
\newblock In {\em Proceedings of the IEEE/CVF International Conference on
  Computer Vision}, pages 2821--2830, 2019.

\bibitem{robicquet2016learning}
Alexandre Robicquet, Amir Sadeghian, Alexandre Alahi, and Silvio Savarese.
\newblock Learning social etiquette: Human trajectory understanding in crowded
  scenes.
\newblock In {\em European conference on computer vision}, pages 549--565,
  2016.

\bibitem{salzmann2020trajectron++}
Tim Salzmann, Boris Ivanovic, Punarjay Chakravarty, and Marco Pavone.
\newblock Trajectron++: Dynamically-feasible trajectory forecasting with
  heterogeneous data.
\newblock {\em arXiv preprint arXiv:2001.03093}, 2020.

\bibitem{sun2020waymo_open}
Pei Sun, Henrik Kretzschmar, Xerxes Dotiwalla, Aurelien Chouard, Vijaysai
  Patnaik, Paul Tsui, James Guo, Yin Zhou, Yuning Chai, Benjamin Caine, et~al.
\newblock Scalability in perception for autonomous driving: Waymo open dataset.
\newblock In {\em Proceedings of the IEEE/CVF Conference on Computer Vision and
  Pattern Recognition}, pages 2446--2454, 2020.

\bibitem{suo2021trafficsim}
Simon Suo, Sebastian Regalado, Sergio Casas, and Raquel Urtasun.
\newblock Trafficsim: Learning to simulate realistic multi-agent behaviors.
\newblock In {\em Conference on Computer Vision and Pattern Recognition
  (CVPR)}, 2021.

\bibitem{tang2019multiple}
Charlie Tang and Russ~R Salakhutdinov.
\newblock Multiple futures prediction.
\newblock In {\em {NeurIPS}}, 2019.

\bibitem{thiede2019analyzing}
Luca~Anthony Thiede and Pratik~Prabhanjan Brahma.
\newblock Analyzing the variety loss in the context of probabilistic trajectory
  prediction.
\newblock In {\em Proceedings of the IEEE/CVF International Conference on
  Computer Vision}, pages 9954--9963, 2019.

\bibitem{kate_cbp}
Ekaterina Tolstaya, Reza Mahjourian, Carlton Downey, Balakrishnan Vadarajan,
  Benjamin Sapp, and Dragomir Anguelov.
\newblock Identifying driver interactions via conditional behavior prediction.
\newblock {\em 2021 IEEE International Conference on Robotics and Automation
  (ICRA)}, 2021.

\bibitem{REF:yang2021auto4d}
Bin Yang, Min Bai, Ming Liang, Wenyuan Zeng, and Raquel Urtasun.
\newblock Auto4d: Learning to label 4d objects from sequential point clouds,
  2021.

\bibitem{yu2020bdd100k}
Fisher Yu, Haofeng Chen, Xin Wang, Wenqi Xian, Yingying Chen, Fangchen Liu,
  Vashisht Madhavan, and Trevor Darrell.
\newblock Bdd100k: A diverse driving dataset for heterogeneous multitask
  learning.
\newblock In {\em Proceedings of the IEEE/CVF Conference on Computer Vision and
  Pattern Recognition}, pages 2636--2645, 2020.

\bibitem{interactiondataset}
Wei Zhan, Liting Sun, Di Wang, Haojie Shi, Aubrey Clausse, Maximilian Naumann,
  Julius K\"ummerle, Hendrik K\"onigshof, Christoph Stiller, Arnaud de
  La~Fortelle, and Masayoshi Tomizuka.
\newblock {INTERACTION} {Dataset}: {An} {INTERnational}, {Adversarial} and
  {Cooperative} {moTION} {Dataset} in {Interactive} {Driving} {Scenarios} with
  {Semantic} {Maps}.
\newblock {\em arXiv:1910.03088 [cs, eess]}, 2019.

\bibitem{zhao2020tnt}
Hang Zhao, Jiyang Gao, Tian Lan, Chen Sun, Benjamin Sapp, Balakrishnan
  Varadarajan, Yue Shen, Yi Shen, Yuning Chai, Cordelia Schmid, et~al.
\newblock Tnt: Target-driven trajectory prediction.
\newblock {\em arXiv preprint arXiv:2008.08294}, 2020.

\bibitem{zhou2020end}
Yin Zhou, Pei Sun, Yu Zhang, Dragomir Anguelov, Jiyang Gao, Tom Ouyang, James
  Guo, Jiquan Ngiam, and Vijay Vasudevan.
\newblock End-to-end multi-view fusion for 3d object detection in lidar point
  clouds.
\newblock In {\em Conference on Robot Learning}, pages 923--932, 2020.

\end{thebibliography}
}

\clearpage
\appendix

\section{Motion Forecasting Metrics}
Distance error metrics are the most commonly used to compare methods, capturing how close a predicted trajectory (discrete time sequence of states) matches a future object track, under Euclidean distance.  The most common is Average Displacement Error (ADE)~\cite{alahi2016social, pellegrini2009you}.  Because the future is inherently stochastic and multi-modal, most models output a (weighted) set of trajectory hypotheses, and then a minimal error over the set (of constrained size) is reported (i.e. minADE~\cite{chang2019argoverse}). For methods that provide explicit or implicit future probability distributions, the likelihood of the ground truth future trajectory can be used as a metric~\cite{chai2019multipath, salzmann2020trajectron++, rhinehart2018r2p2, rhinehart2019precog}.  Framing the problem instead as one of detection of future locations, Argoverse~\cite{chang2019argoverse} employs Miss Rate within 2 meters as their primary metric, which has the benefit to being tolerant to outliers. A number of metrics including minADE have been extended for use with jointly predicted agent trajectories\cite{casas2020implicit}.

\section{Dataset Splits}
The dataset provides 6 different splits of the original set of 20 second scenarios. The scenarios are first split into training, validation and test sets. This is done by hashing a string containing the date of the data capture and the unique ID of the vehicle used to capture the data. The hashed values are split into mutually exclusive 70\% training, 15\% validation, and 15\% testing subsets of the 20 second scenarios. From these 3 subsets we generate examples by extracting 9.1 second windows from the longer 20 second scenarios. Each 9.1 second window contains 91 time steps at 10Hz - 10 history samples, 1 sample at the current time, and 80 future steps. We extract 5 different sets of windowed examples from the respective 20 second splits, training, validation, testing, validation interactive, and testing interactive. The training set contains 9.1 second windows starting at times \{0, 2, 4, 5, 6, 8, 10\} seconds within the 20 second scenarios. The validation and testing sets contain 9.1 second windows starting at times \{0, 5, 10\} seconds. The validation interactive and testing interactive sets contain 9 second windows starting at times \{4, 5, 6\} seconds to focus on the interactive portion of the scenario. The 5 windowed sets are included in the published dataset along with the full 20 second training set. Each of the windowed sets contains a list of objects in the scene to be predicted. The training, validation, and testing sets contain up to 8 objects per scenario chosen to include at least 2 objects of each type if available. Selection is biased to include objects that do not follow a constant velocity model or straight paths. For the validation interactive and testing interactive sets, only the mined interactive agent pair objects are included in the list of objects to predict. In addition, each object to predict has a difficulty level based on how easily it is predicted by an LSTM extrapolation model.

\section{Metrics Details}
\label{sec:Metrics Details}

\PAR{Overlap rate (OR) details.} A binary indicator is assigned to each sample alerting of self-overlapping. The average over the dataset creates the overlap rate. We only consider the highest scoring joint prediction $\bptilde$ here.
Our metric counts an overlap with the following criteria: given the joint predicted trajectories of $A$ agents, an overlap is counted if the rotated bounding box of any of the $A$ agents overlaps with any other visible object at  any time step within the prediction interval $T$. Note that agents not visible at prediction time (due to their later appearance) are not considered for potential overlaps.
Consider $\setG_t = \{\tilde{s}_{a,t} \forall a, g_{b,t} \forall b \in 1 \dots B\}$ where $\tilde{s}_{a,t}$ are waypoints from $\bptilde$ at time $t$, and $g_{b,t}$ are groundtruth waypoints from $B$ nearby environmental agents, the single overlap indicator is defined as:
\begin{equation} 
\mu_{\mbox{OR}}(\bme) = \sum_t \sum_a \sum_{s' \in {\setG_t \setminus \tilde{s}_{a,t}}}
\mathbbm{1} [\mbox{IOU}(b(\tilde{s}_{a, t}), b(s'_t)) > 0]
\label{eq:mr}
\end{equation}
where $b(.)$ is a function to derive a 5-dof ($x$, $y$, width, length and heading) bounding box from a waypoint. The groundtruth bounding box is used for an environmental agent. For a predicted waypoint $s_{a,t}$, we derive the heading from the derivative to the previous waypoint and use the groundtruth bounding box sizes. \mbox{$IOU(\cdot)$} computes the intersection-over-union between two 5-dof boxes.

\PAR{Miss rate (MR) details.} The indicator function $f(.)$ in \eqref{eq:mr} is defined as follows:
\begin{align}
\label{eq:f}
f(.) &= \mathbbm{1}[x_a^k > \lambda^{lon}] \vee  \mathbbm{1}[y_a^k > \lambda^{lat}] \\
[x_a^k, y_a^k] &\coloneqq (\hat{s}_a - s_a^k) \cdot \bR_a \nonumber
\end{align}
where $\bR_a$ is a 2D rotation matrix defined by the heading of agent $a$ at the timestamp $0$. $\lambda^{lon}$ and $\lambda^{lat}$ are longitudinal and lateral thresholds. Since agents can have different speed at time $0$, we scale these thresholds by their speed so that we do not over-penalize faster agents: $\lambda^{lon} = \lambda^{lon}_0 \gamma(v_x)$ and $\lambda^{lat} = \lambda^{lat}_0 \gamma(v_y)$, where $\gamma(v) = (\max(0, \min(1, (v - \upsilon_L) / (\upsilon_H - \upsilon_L))) / 2 + 0.5$. We set $\upsilon_H$ to 11 m/s and $\upsilon_L$ to 1.4 m/s. The thresholds dependent on $T$ are as follows:
\begin{center}
\begin{tabular}{| c | c | c |}
\hline
  & $\lambda^{lat}_0$ & $\lambda^{lon}_0$ \\ 
 \hline
 T=3 seconds & 1 & 2 \\
 \hline
 T=5 seconds & 1.8 & 3.6 \\   
 \hline
 T=8 seconds & 3 & 6 \\
 \hline
\end{tabular}
\end{center}

\section{Overlap Metric}
We use a marginal overlap-based metric with the simple baseline models to quantify the difficulty and interactivity in our dataset.  We consider a trajectory for an agent to contain an overlap if at any time point, the agent bounding box overlaps with a ground-truth box at that time. The overlap rate is the number of agents whose trajectories have overlaps divided by the total number of predicted agents.  

\begin{table}[t]
    \centering
    \footnotesize
    \bgroup
    \def\arraystretch{1.05}%
    \begin{tabular}{ll|ccc}
    \toprule
    & & \multicolumn{3}{c}{Overlap Rate} \\
    Val. set & Model & Vehicle & Pedestrian & Cyclist \\
    \midrule
    \multirow{3}{*}{Regular} & Const. Vel. & 38.4\% & 29.8\% & 22.3\% \\
    \cmidrule{2-5}
    & LSTM & 27.9\% & 22.9\% & 22.1\% \\
    \midrule
    \multirow{3}{*}{Interactive} & Const. Vel. & 44.2\% & 30.6\% & 27.0\%  \\
    \cmidrule{2-5}
    & LSTM & 36.3\% & 32.3\% & 25.6\% \\
    \bottomrule
    \end{tabular}%
    \egroup
    \caption{\textbf{The interactive split of the data has more overlaps per scene.} Despite the interactive set only requiring predictions for two agents instead of up to eight agents for the regular dataset, the split contains more scenes where a constant velocity model or an LSTM model -- neither of which models other agents -- produces at least one overlap. Statistics are reported on the validation set for both dataset splits. The marginal-based overlap metric is used for both splits so that the rates can be compared across the splits. Constant velocity model only predicts a single trajectory per agent. For the LSTM model, the highest scoring trajectory for each agent is used.}
    \label{tab:split_stats}
\end{table}

We compute the overlap rate for the constant velocity model and compare the performance between the regular split and interactive split of the dataset. For the constant velocity model, we found that 38.4\% of predicted vehicles in the regular split, and 44.2\% of predicted vehicles in the interactive split have trajectories that overlap with a ground-truth (Table \ref{tab:split_stats}).  This shows that the interactive split is more challenging, and suggests that more interactions between agents in that split.

\section{Conditional Model Details} \label{cbp_appendix}

\begin{figure}[t]
    \centering
    \includegraphics[width=0.9\linewidth]{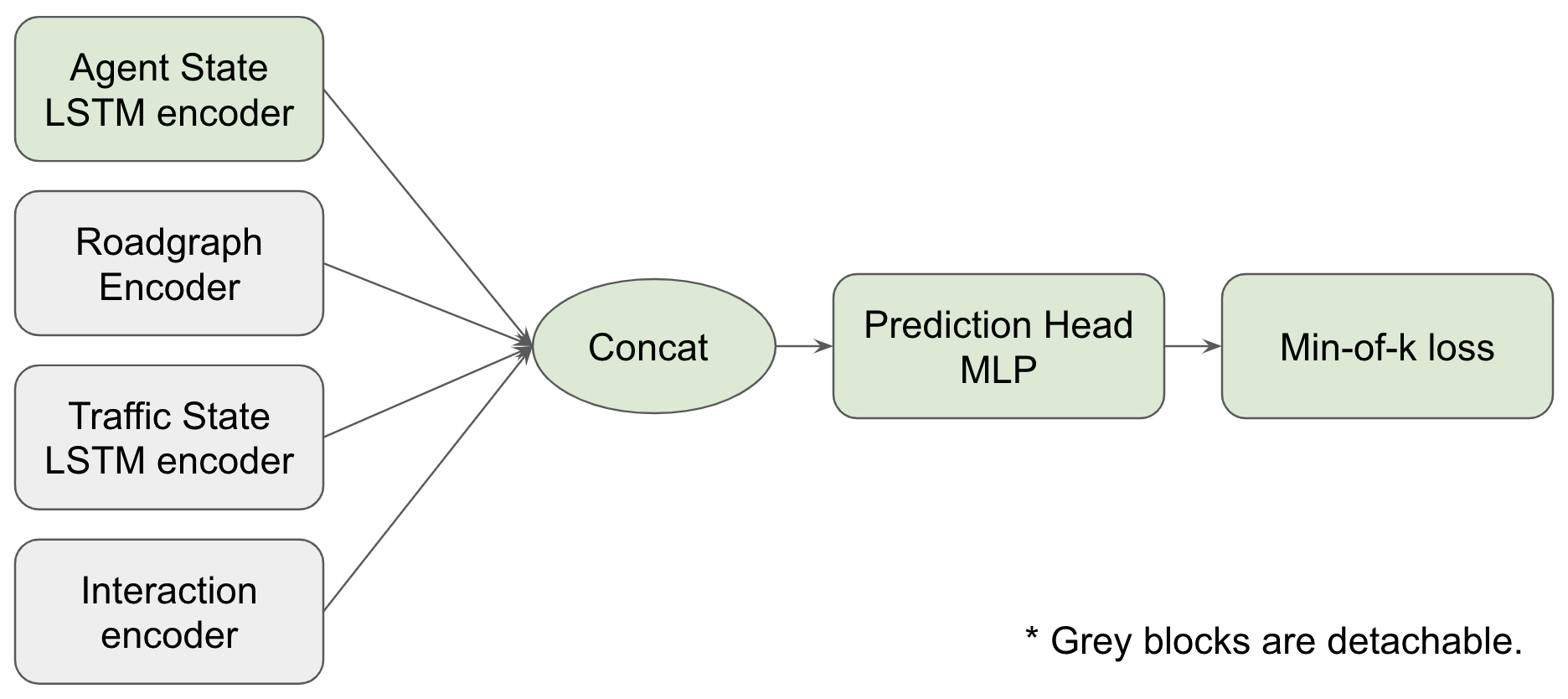}
    \hspace{0.1cm}
    \caption{\textbf{Diagram of baseline architecture.} An illustration of the baseline architecture employed for the family of learned models with a base LSTM encoder for agent states. The three detachable components are a roadgraph polyline encoder \cite{gao2020vectornet}, a traffic state LSTM encoder, and a high-order interactions encoder following \cite{gao2020vectornet}. The trajectories are predicted through a MLP with min-of-$k$ loss.}
    \label{fig:baseline_architectures}
\end{figure}

The model we use for conditional behavior prediction is based on the baseline model we describe in \ref{ssec: baseline}. Figure \ref{fig:baseline_architectures} provides an overview diagram of the proposed model. We use the LSTM encoder and all three enhancements (roadgraph encoding with polylines , traffic signal states encoded in an LSTM, modeling high-order interactions with a global interaction graph). To make this model suitable for conditional predictions, we add an early fusion conditional encoder similar to \cite{kate_cbp}. Just like \cite{kate_cbp}, we train the model to do both conditional and unconditional prediction by passing in a randomly selected query agent's ground truth future trajectory as conditional query input in 95\% of training samples while providing no conditional query in the other 5\%. We generate 6 predictions per agent and evaluate the KL divergence over the full 8 second future trajectory.

\section{Videos}
The included videos show visualization of some samples of scenarios from the dataset including those in Figure~\ref{fig:yield} and Figure~\ref{fig:accelerate}.

\begin{figure*}[t]
    \centering
    \includegraphics[width=0.2\linewidth]{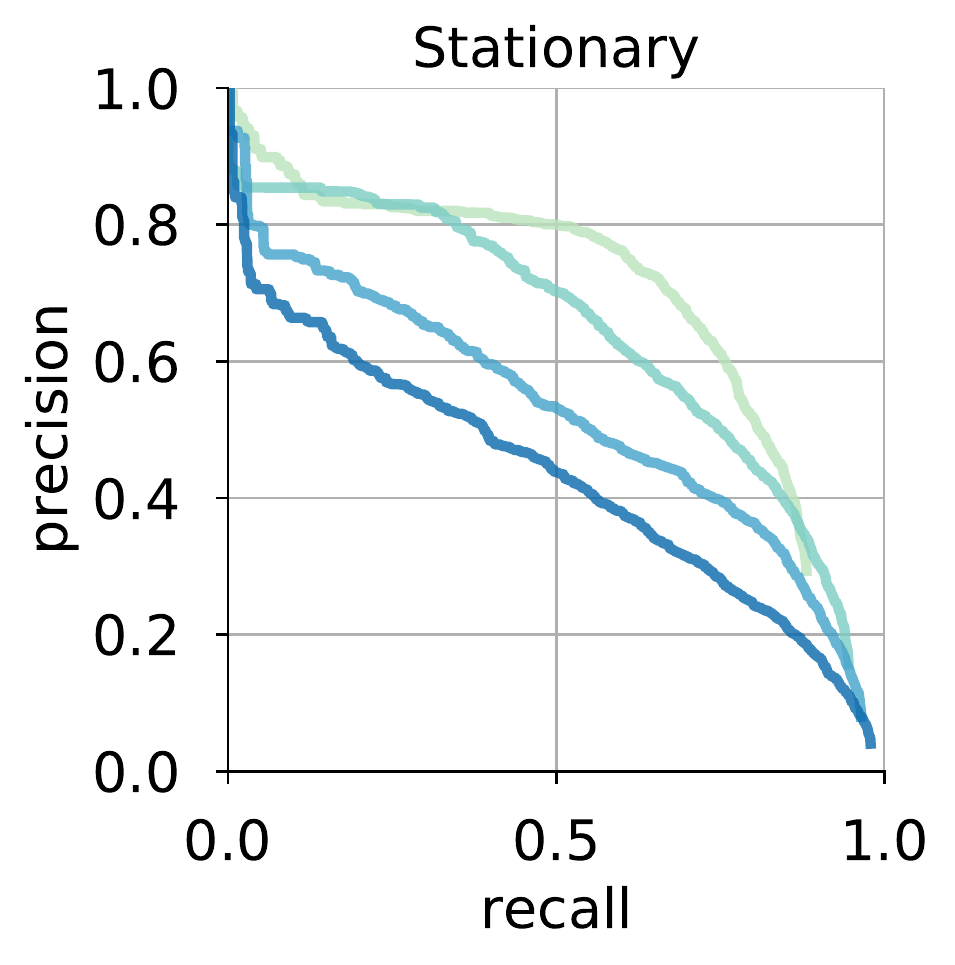}     \includegraphics[width=0.2\linewidth]{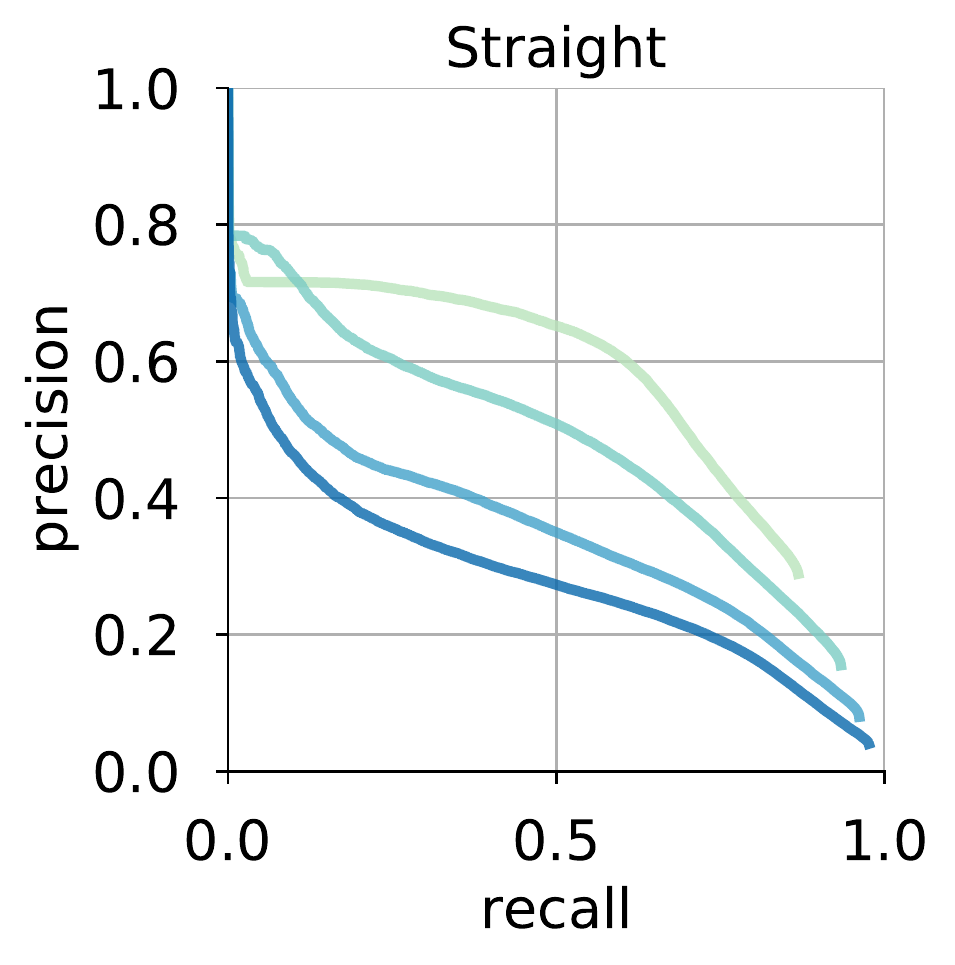}   \includegraphics[width=0.2\linewidth]{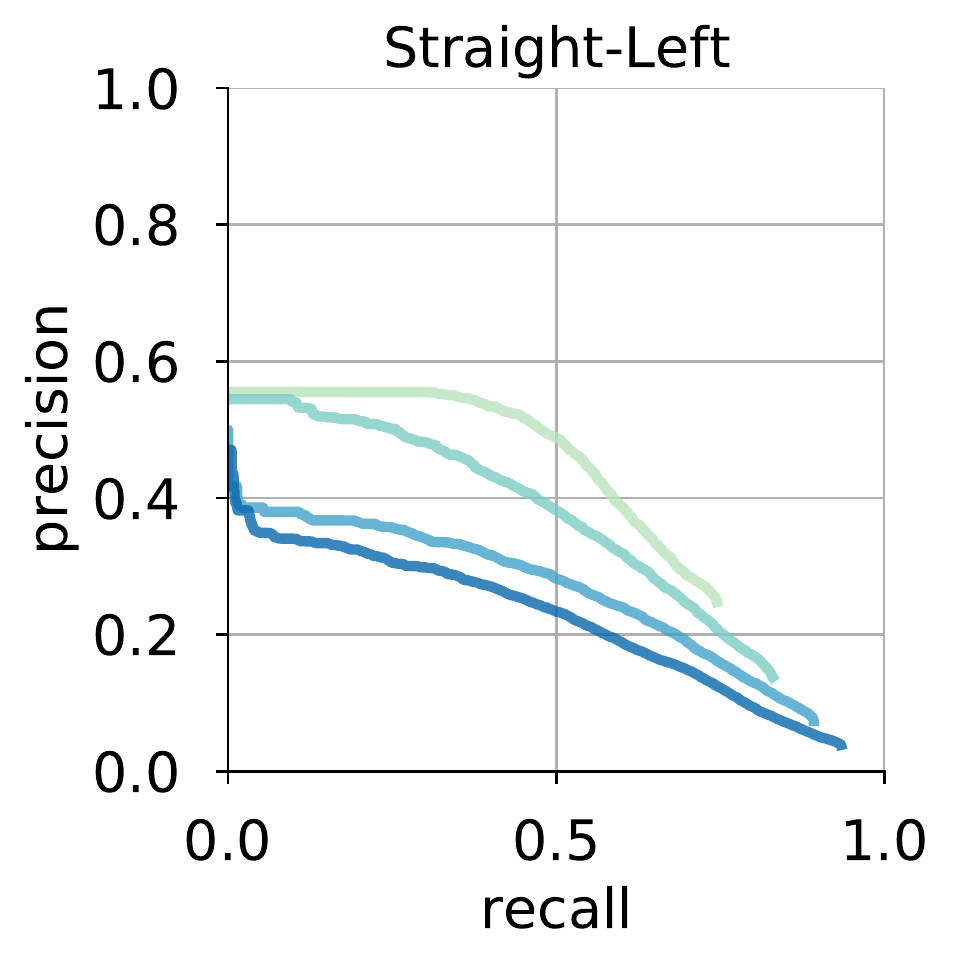}    \includegraphics[width=0.2\linewidth]{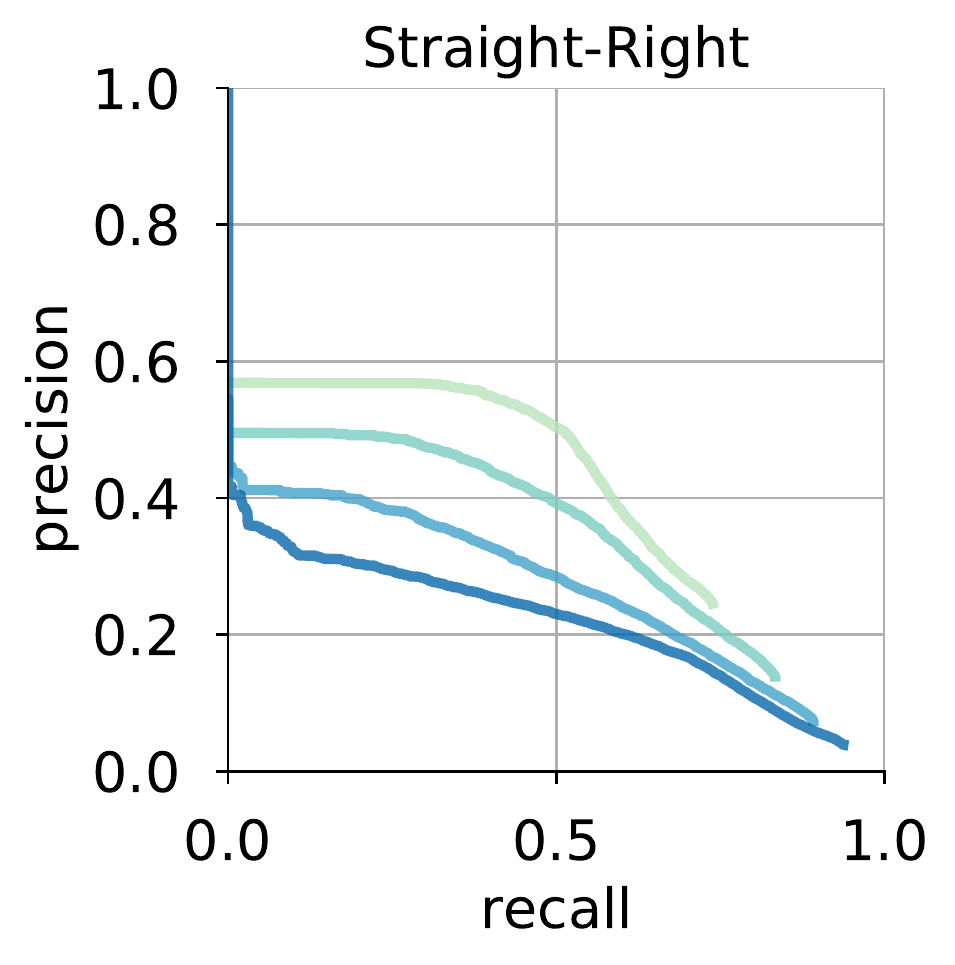} \\
    \includegraphics[width=0.2\linewidth]{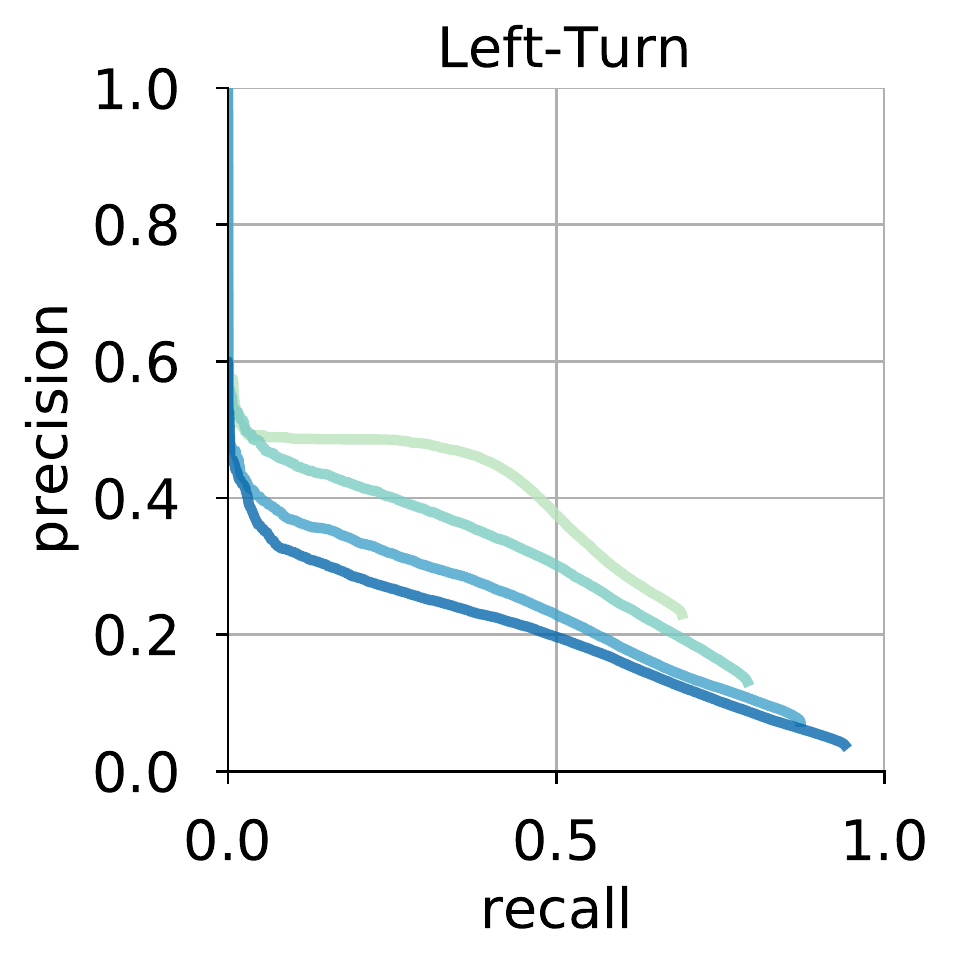} 
    \includegraphics[width=0.2\linewidth]{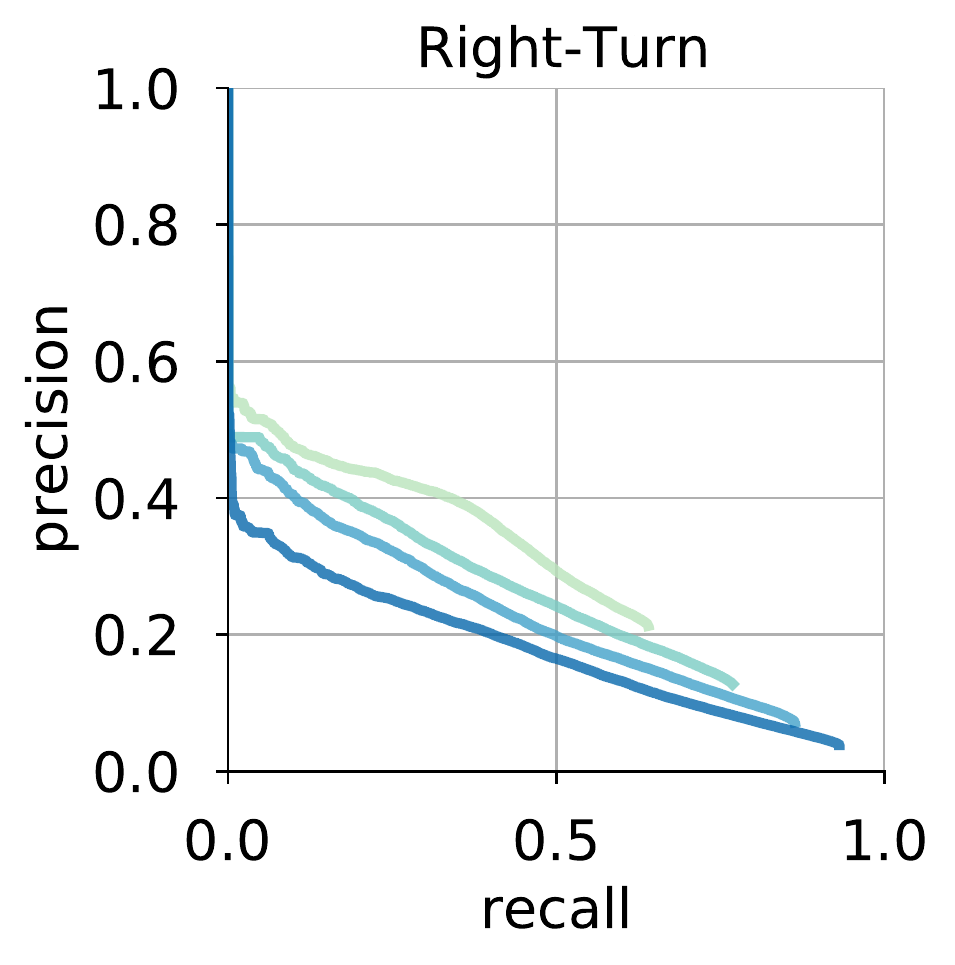} 
    \includegraphics[width=0.2\linewidth]{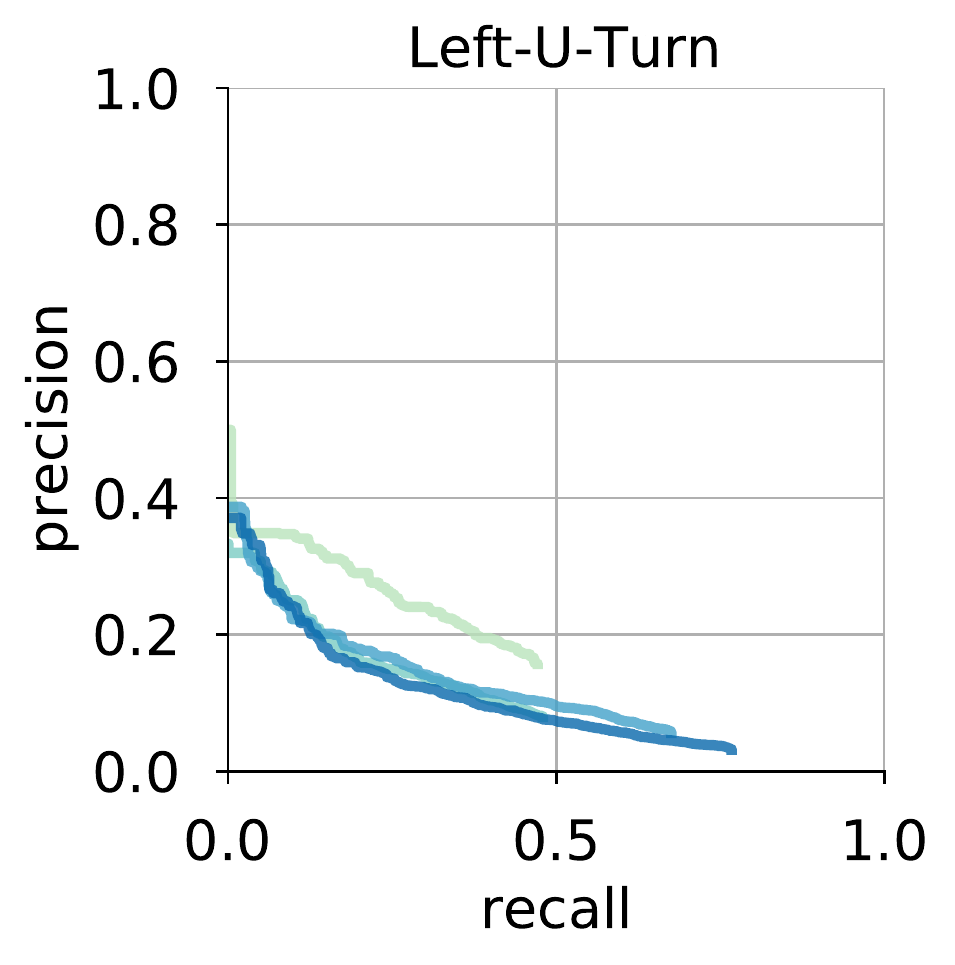}    
    \hspace{0.1cm}
    \includegraphics[width=0.07\linewidth]{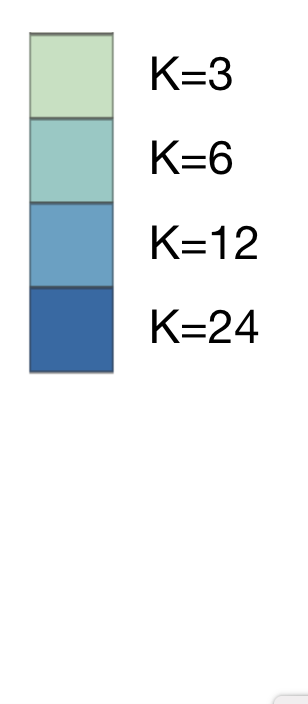} 
    \caption{Precision versus recall curves for increasing number of predictions ($K$) for the polyline model at \textbf{3 seconds} for vehicles across trajectory shape buckets for the standard validation dataset. Recall increases with K but AUC decreases.}
    \label{fig:pr_curves_3sec}
\end{figure*}

\begin{figure*}[t]
    \centering
    \includegraphics[width=0.2\linewidth]{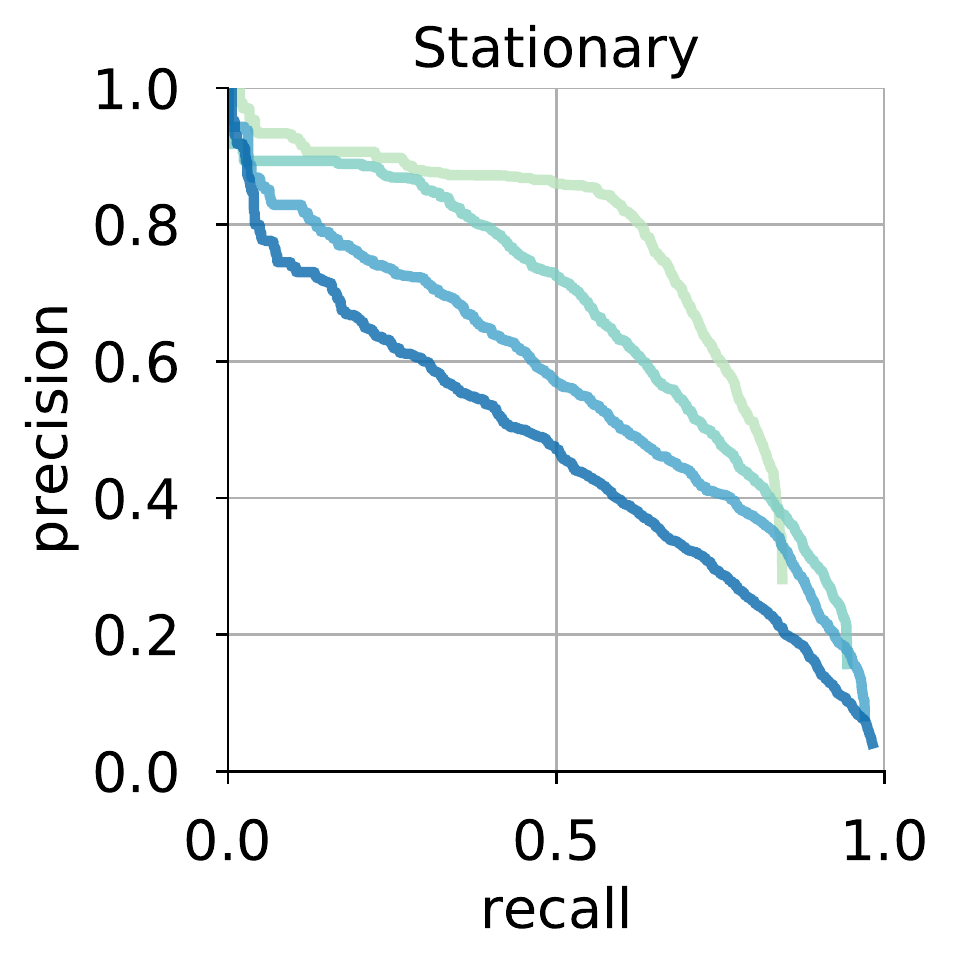}     \includegraphics[width=0.2\linewidth]{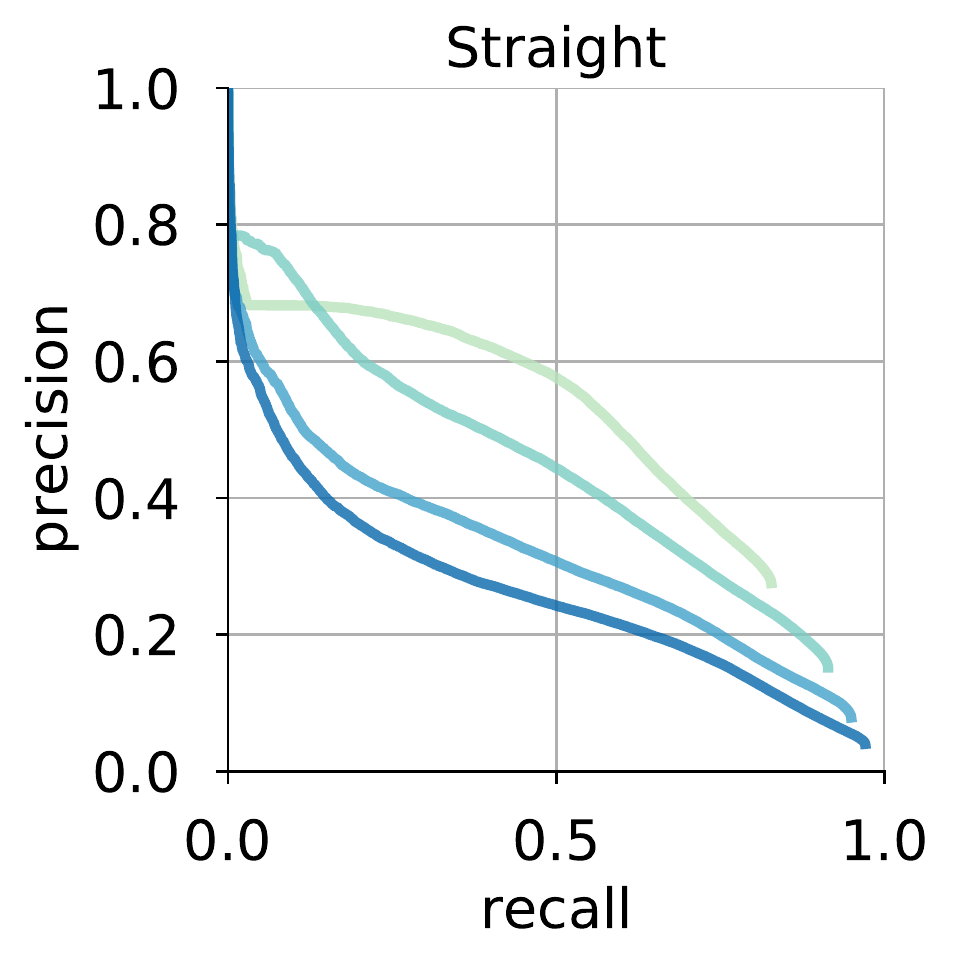}   \includegraphics[width=0.2\linewidth]{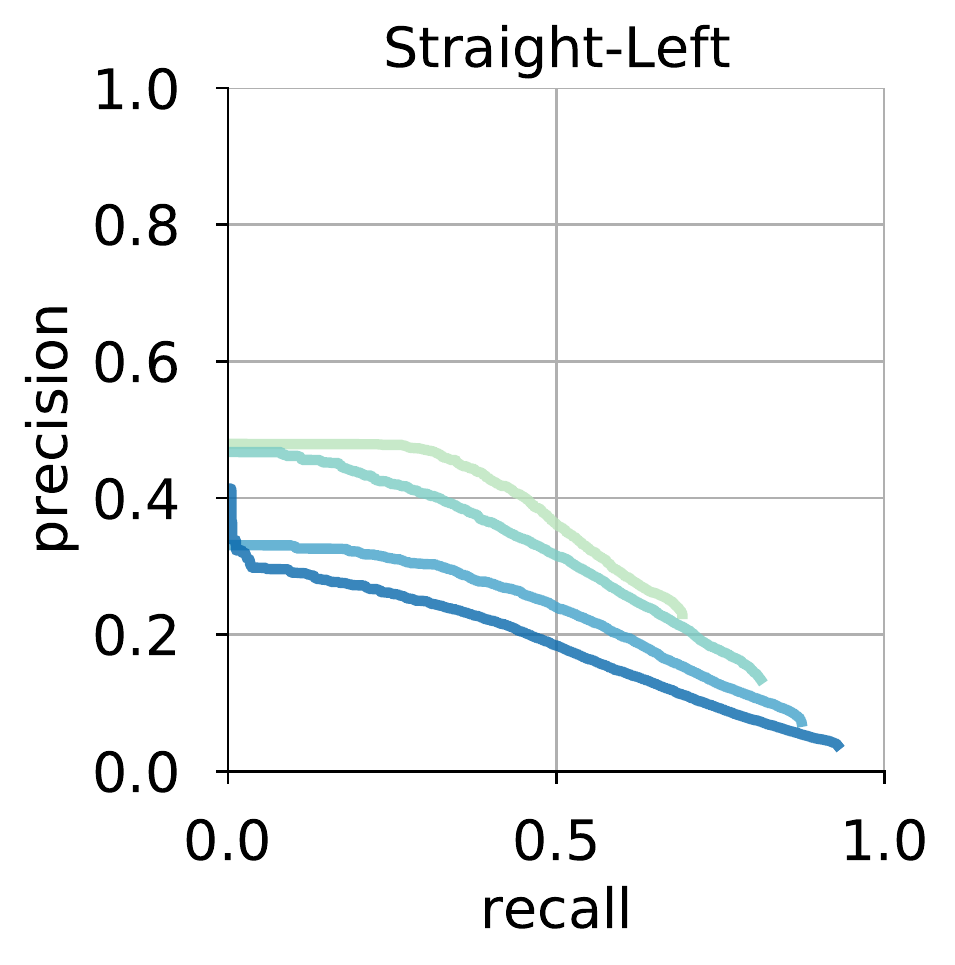}    \includegraphics[width=0.2\linewidth]{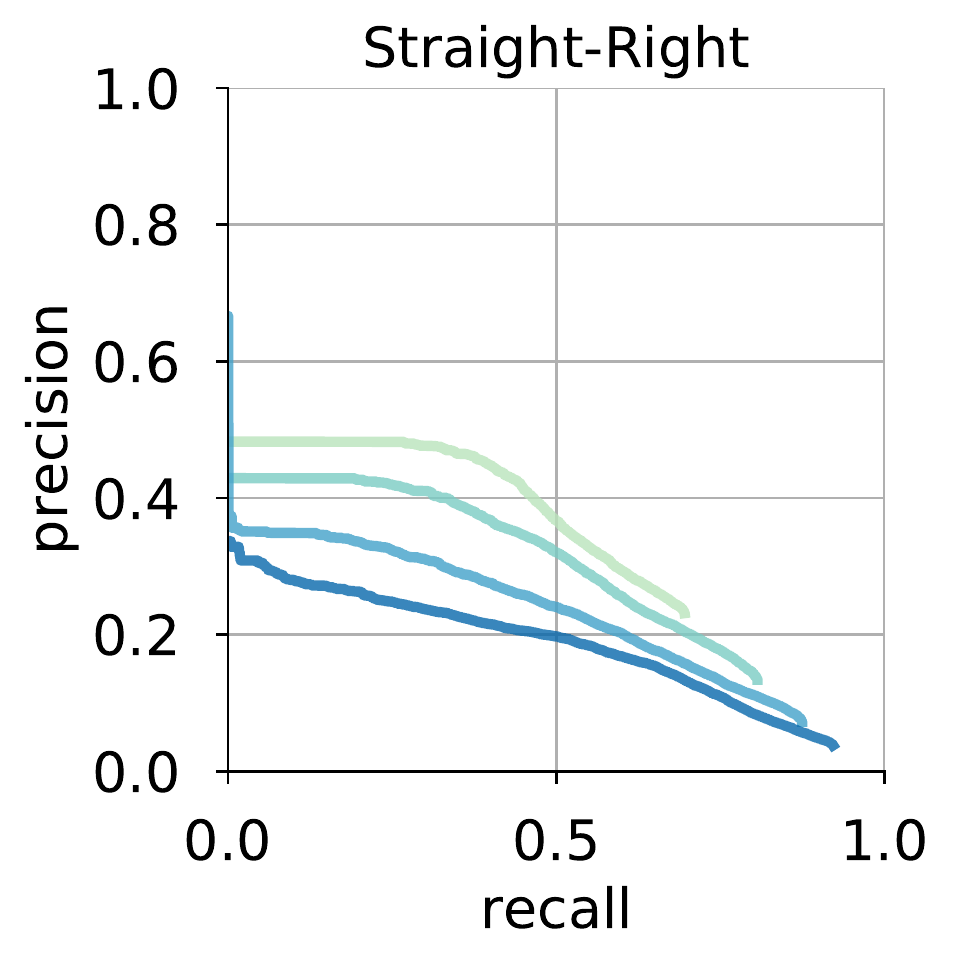} \\
    \includegraphics[width=0.2\linewidth]{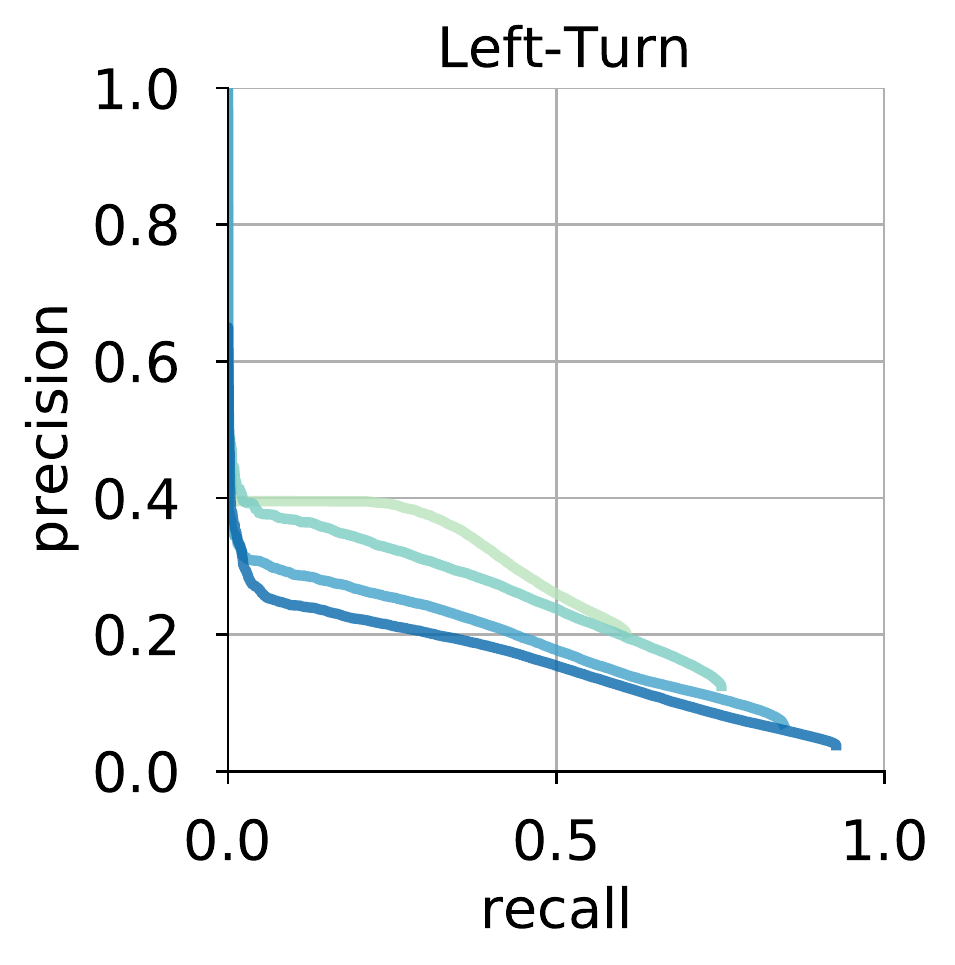} 
    \includegraphics[width=0.2\linewidth]{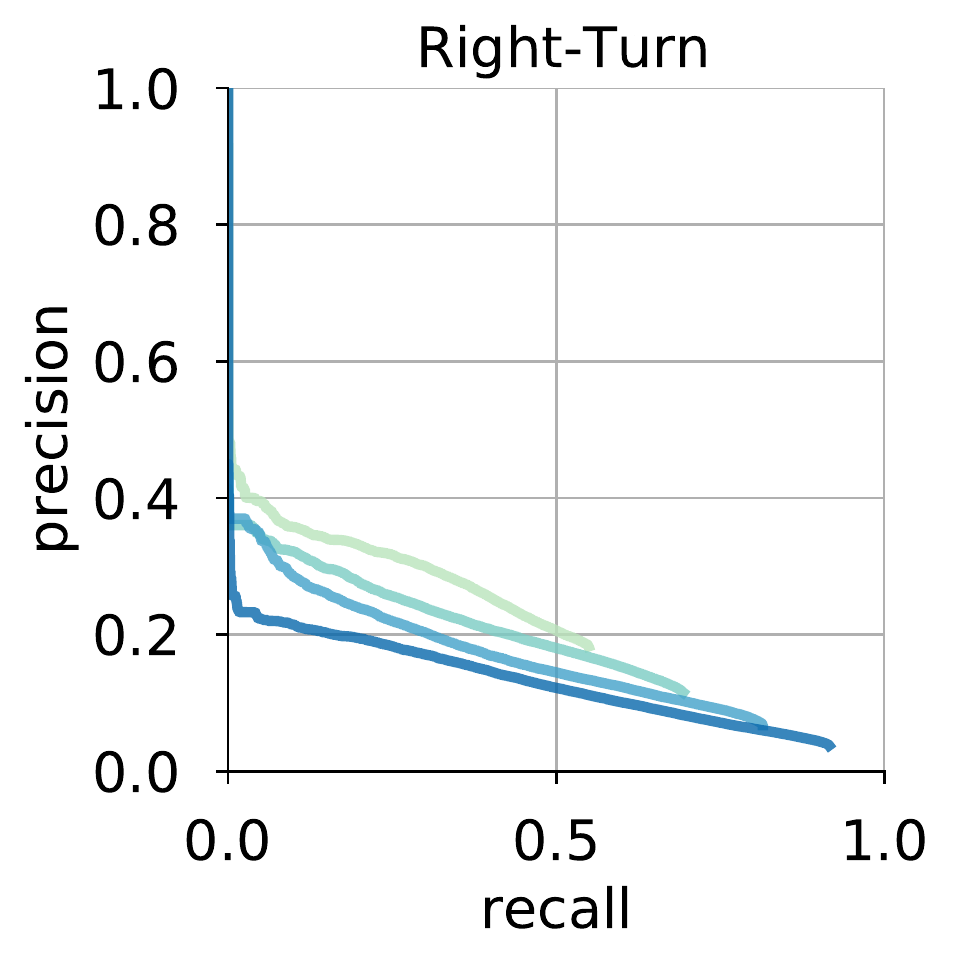} 
    \includegraphics[width=0.2\linewidth]{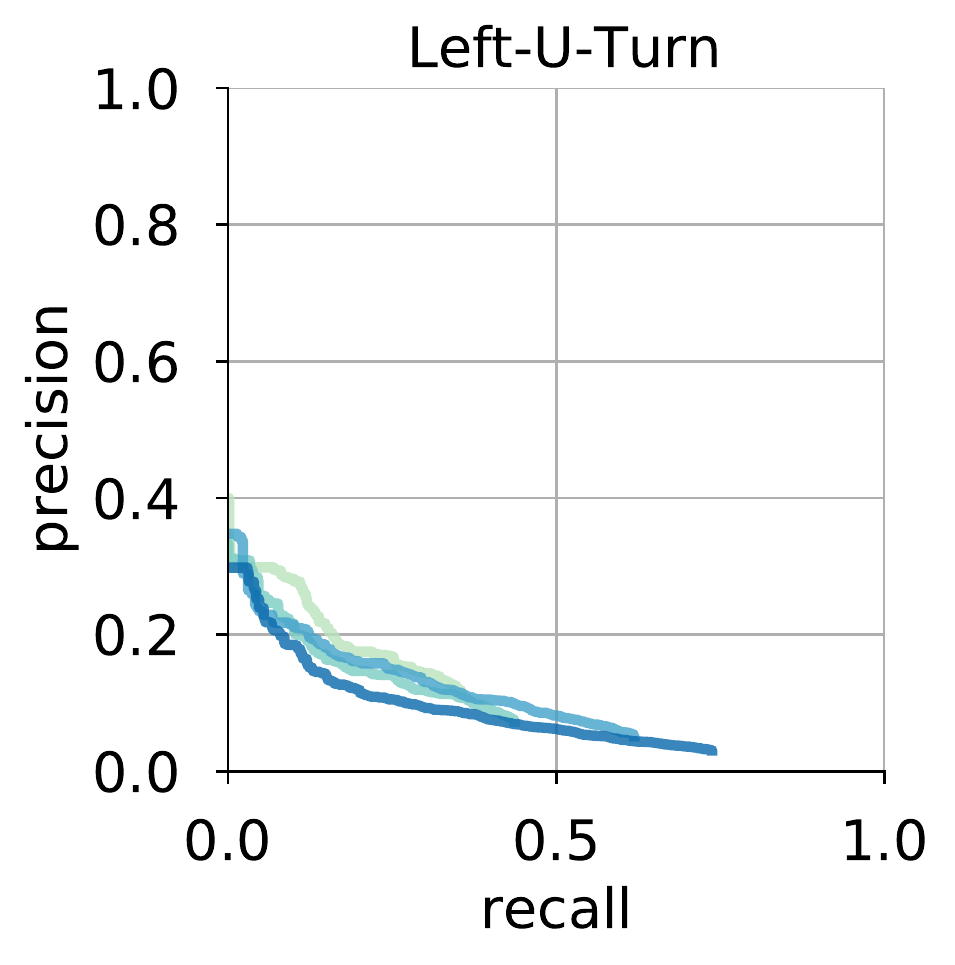}    
    \hspace{0.1cm}
    \includegraphics[width=0.07\linewidth]{figures/PR/WOD-PR-Legend.png} 
    \caption{Precision versus recall curves for increasing number of predictions ($K$) for the polyline model at \textbf{5 seconds} for vehicles across trajectory shape buckets for the standard validation dataset.}
    \label{fig:pr_curves_5sec}
\end{figure*}

\begin{figure*}[t]
    \centering
    \includegraphics[width=0.2\linewidth]{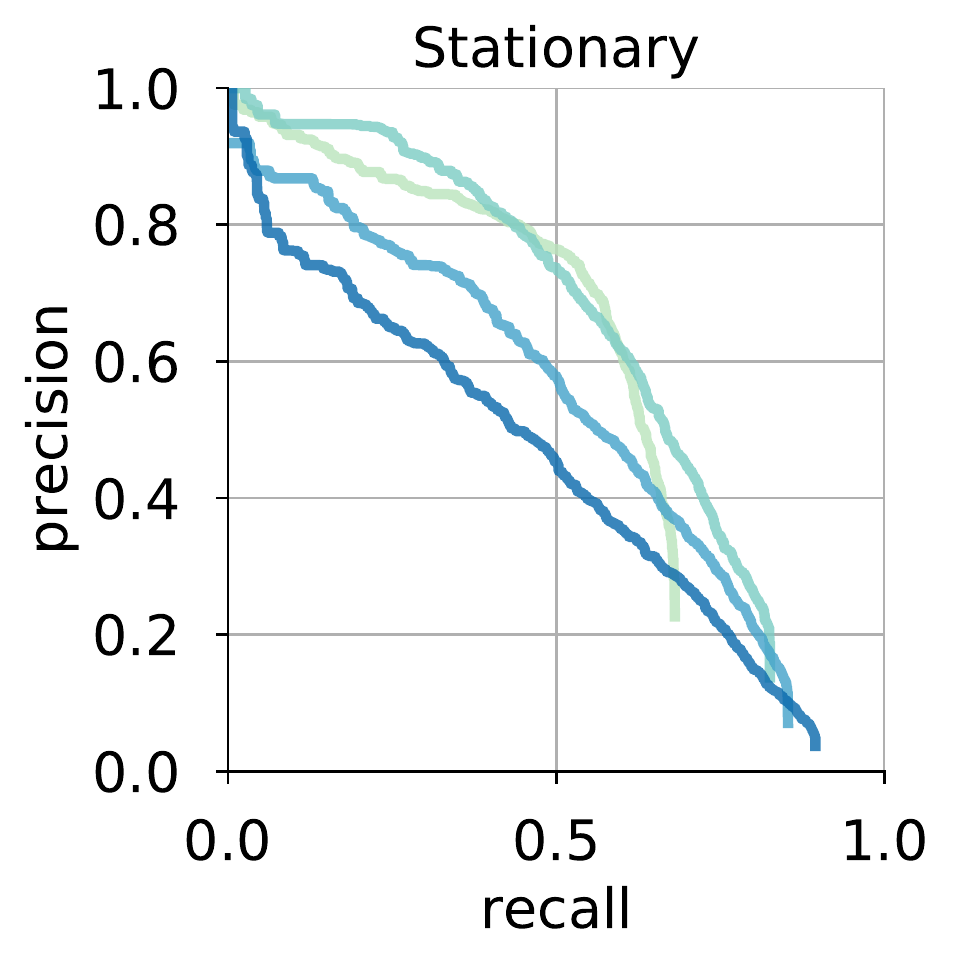}     \includegraphics[width=0.2\linewidth]{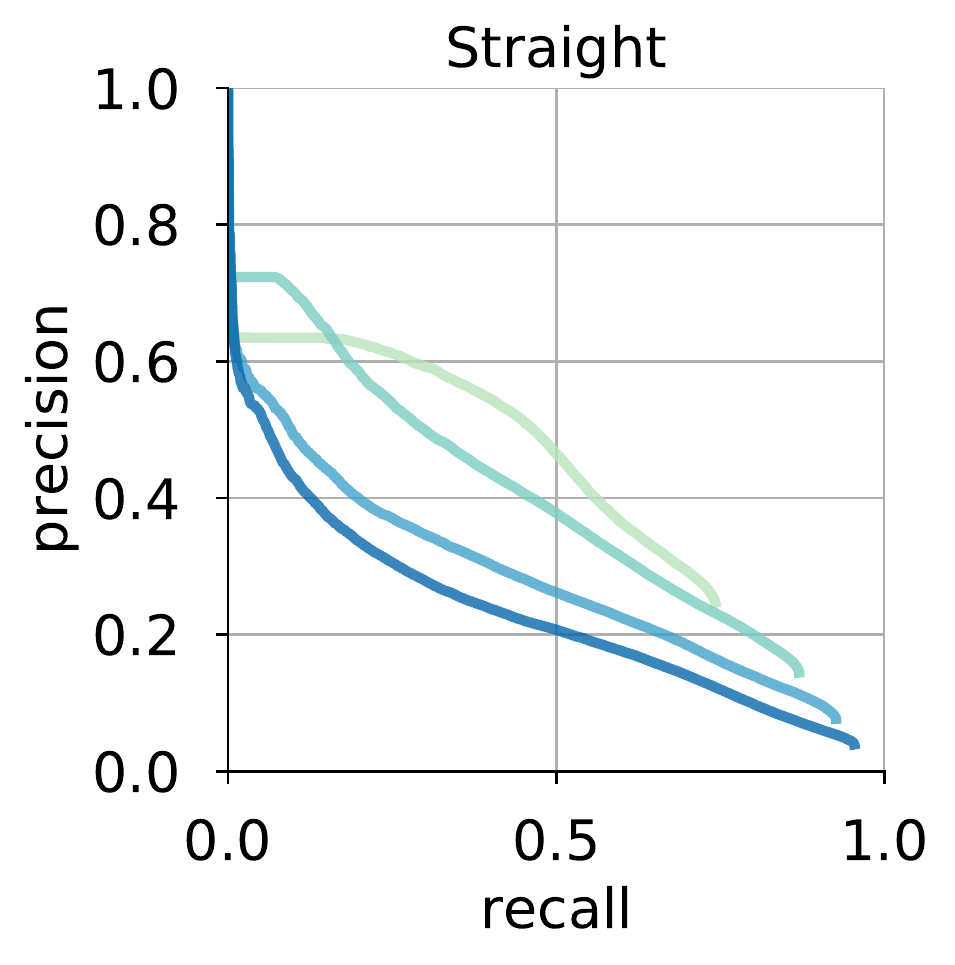}   \includegraphics[width=0.2\linewidth]{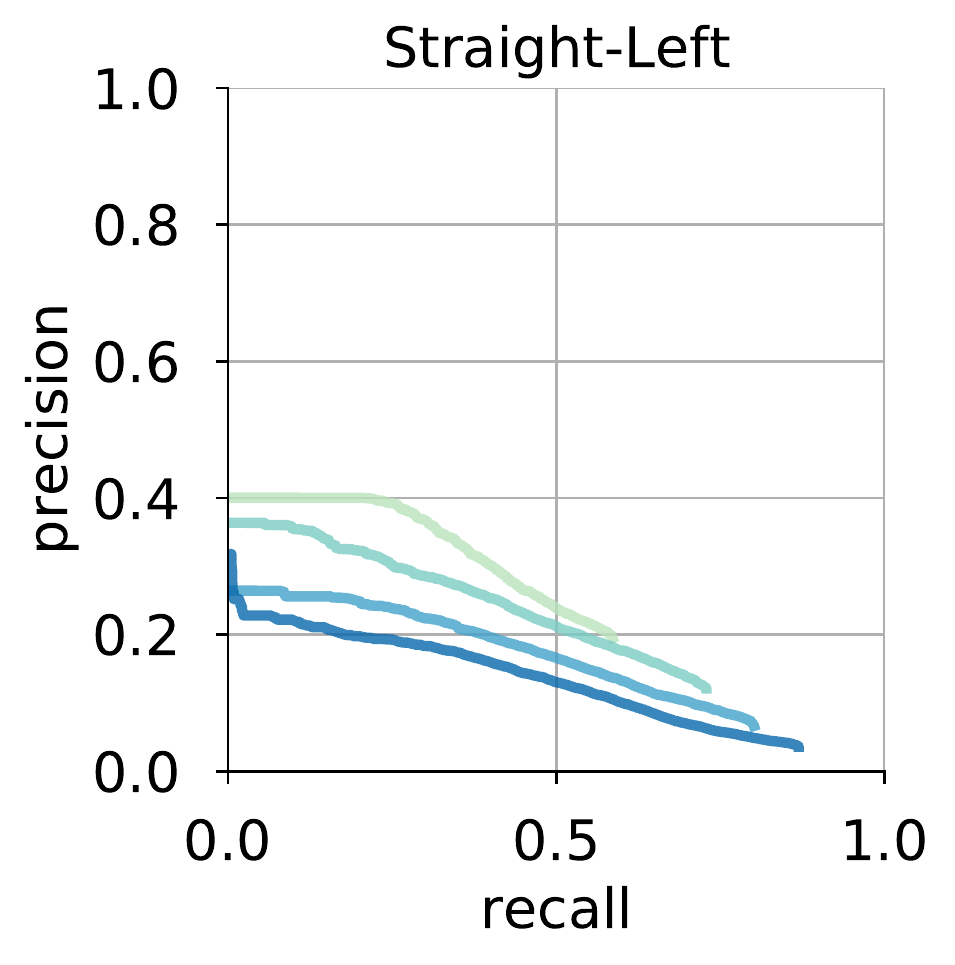}    \includegraphics[width=0.2\linewidth]{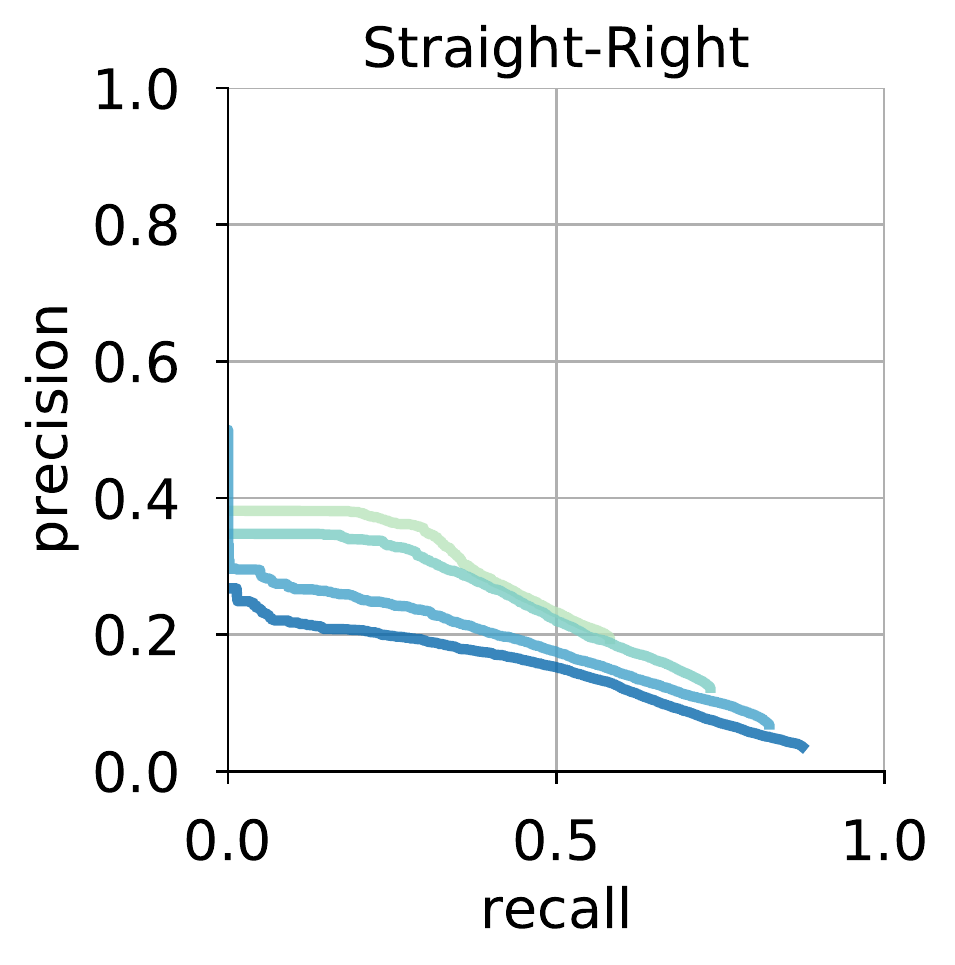} \\
    \includegraphics[width=0.2\linewidth]{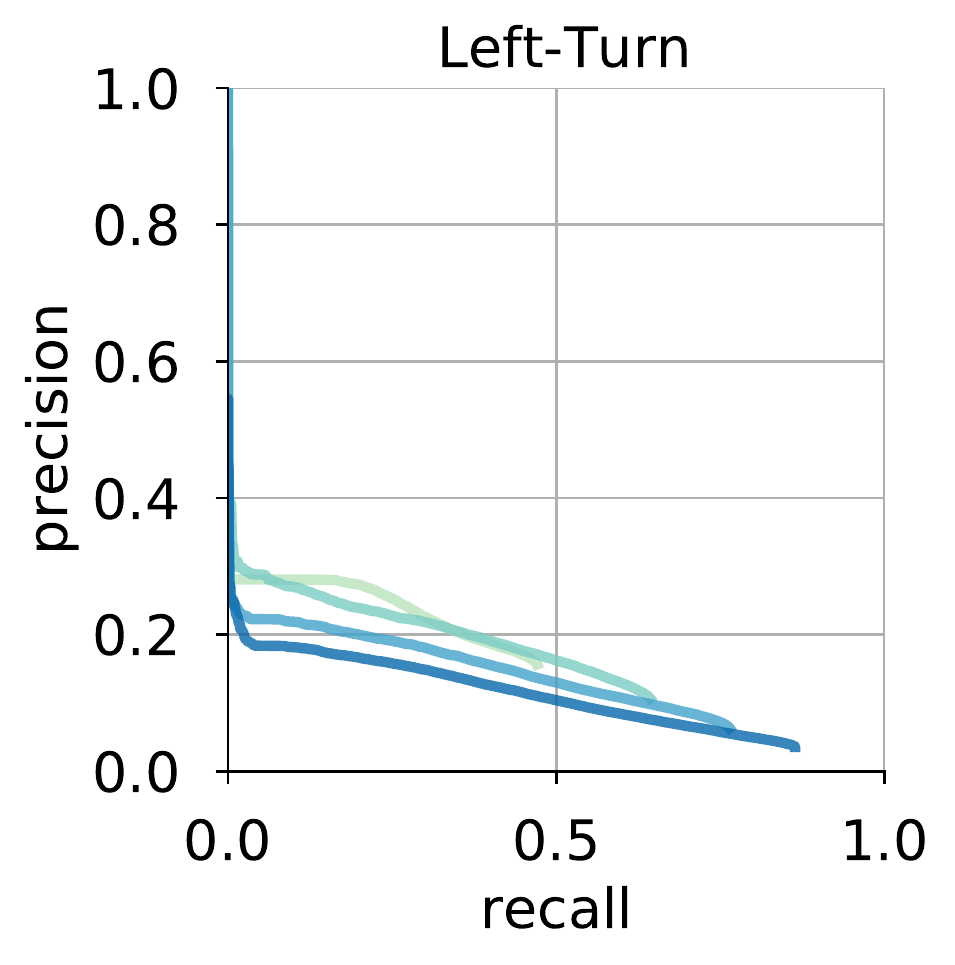} 
    \includegraphics[width=0.2\linewidth]{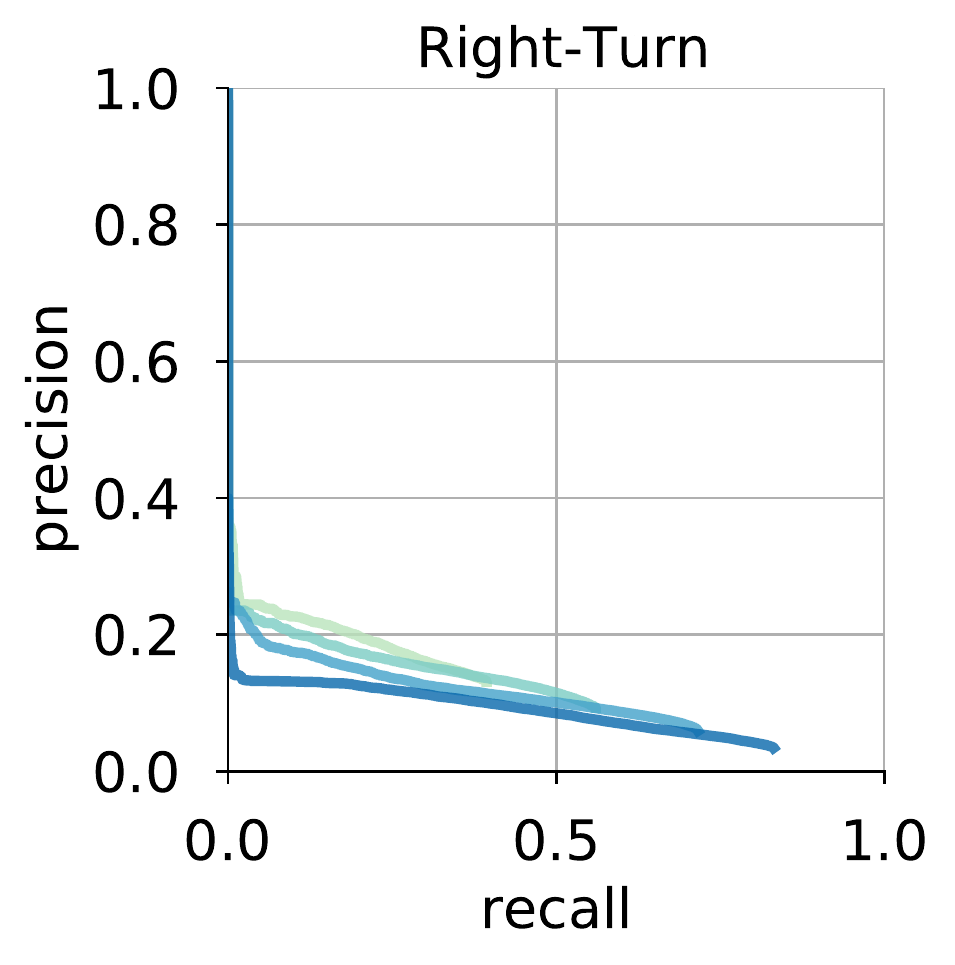} 
    \includegraphics[width=0.2\linewidth]{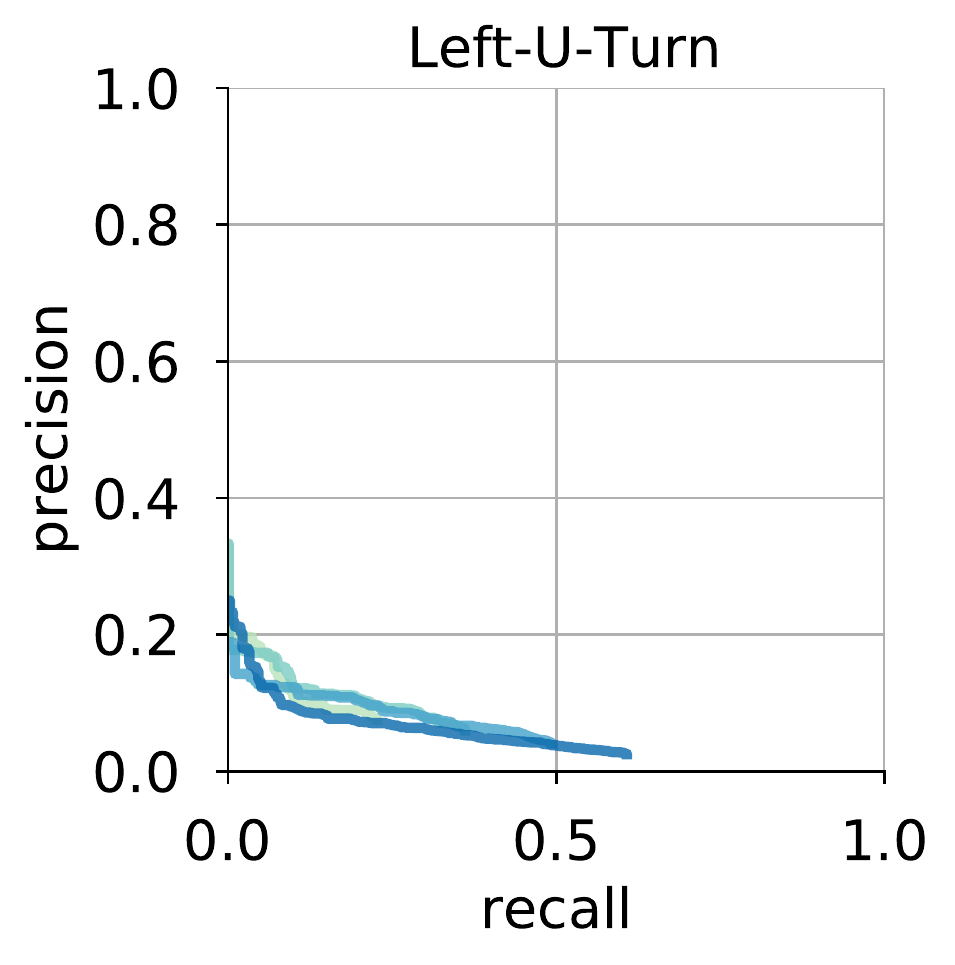}    
    \hspace{0.1cm}
    \includegraphics[width=0.07\linewidth]{figures/PR/WOD-PR-Legend.png} 
    \caption{Precision versus recall curves for increasing number of predictions ($K$) for the polyline model at \textbf{8 seconds} for vehicles across trajectory shape buckets for the standard validation dataset.}
    \label{fig:pr_curves_8sec}
\end{figure*}

\end{document}